\documentclass[journal]{IEEEtran}

\usepackage{graphicx}
\usepackage{subfigure}
\usepackage{lipsum}
\usepackage{cite}
\usepackage{bm}
\usepackage{esvect}
\usepackage{amsmath}
\usepackage[ruled,vlined]{algorithm2e}

\graphicspath{
    {./fig/}
}

\usepackage{color}

\usepackage[normalem]{ulem}

\ifCLASSINFOpdf
\else
\fi

\begin{document}
\title{General Support-Effective Decomposition for Multi-Directional 3D Printing}

\author{Chenming~Wu,
        Chengkai~Dai,
        Guoxin~Fang,
        Yong-Jin~Liu,
        and~Charlie~C.L.~Wang$^{*}$
\thanks{Manuscript received March 16, 2019; revised June 18, 2019; accepted August 8, 2019.}
\thanks{C. Wu and Y.-J. Liu are with Beijing National Research Center for Information Science and Technology, Department of Computer Science and Technology, Tsinghua University, Beijing, China.}
\thanks{C. Dai and G. Fang are with the Department of Design Engineering, Delft University of Technology, Netherlands. Part of this work was completed when they worked at the Chinese University of Hong Kong.}
\thanks{C.C.L. Wang is with the Department of Mechanical and Automation Engineering, The Chinese University of Hong Kong, Shatin, Hong Kong.}
\thanks{$^{*}$Corresponding Authors (cwang@mae.cuhk.edu.hk)}
}

\markboth{}%
{Wu \MakeLowercase{\textit{et al.}}: Support-Effective Multi-Directional Additive Manufacturing}

\maketitle

\begin{abstract}
We present a method for fabricating general models with multi-directional 3D printing systems by printing different model regions along with different directions. The core of our method is a support-effective volume decomposition algorithm that minimizes the area of the regions with large overhangs. A beam-guided searching algorithm with manufacturing constraints determines the optimal volume decomposition, which is represented by a sequence of clipping planes. While current approaches require manually assembling separate components into a final model, our algorithm allows for directly printing the final model in a single pass. It can also be applied to models with loops and handles. A supplementary algorithm generates special supporting structures for models where supporting structures for large overhangs cannot be eliminated. 
We verify the effectiveness of our method using two hardware systems: a Cartesian-motion based system and an angular-motion based system. A variety of 3D models have been successfully fabricated on these systems. 
\end{abstract}

\textit{Note to Practitioner}
\begin{abstract}
In conventional planar-layer based 3D printing systems, supporting structures need to be added at the bottom of large overhanging regions to prevent material collapse. Supporting structures used in single-material 3D printing technologies have three major problems: being difficult to remove, introducing surface damage, and wasting material. This research introduces a method to improve 3D printing by adding rotation during the manufacturing process. To keep the hardware system relatively inexpensive, the hardware, called a \textit{multi-directional 3D printing system}, only needs to provide unsynchronized rotations. In this system, models are subdivided into different regions, and then the regions are printed in different directions. We develop a general volume decomposition algorithm for effectively reducing the area that needs supporting structures. When supporting structures cannot be eliminated, we provide a supplementary algorithm for generating supports compatible with multi-directional 3D printing. Our method can speed up the process of 3D printing by saving time in producing and removing supports.
\end{abstract}


\IEEEpeerreviewmaketitle

\section{Introduction}\label{secIntroduction}

\begin{figure}[t] 
\centering
\includegraphics[width=\linewidth]{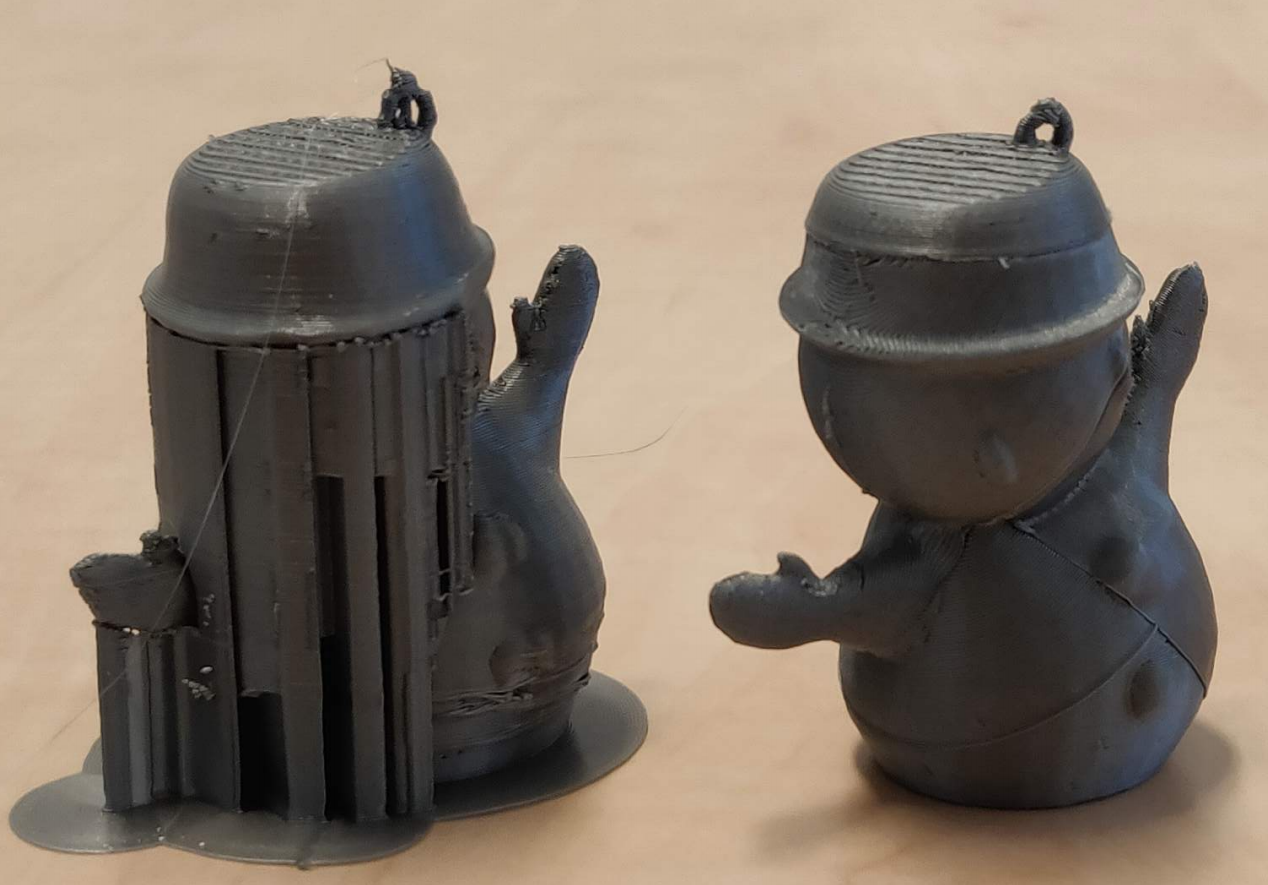}
\caption{Snowman models fabricated with an off-the-shelf FDM printer (left) and our multi-directional printing system that adds only one rotational axis to the same printer (right). By allowing material accumulation along different directions in different regions, our system substantially reduces support necessitation.}
\label{figTeaser}
\end{figure}

\IEEEPARstart{C}{onsumer-grade} \textit{Additive Manufacturing} (AM) devices (3D printers), particularly those based on \textit{Fused Deposition Modeling} (FDM) claim to have the ability to fabricate complex shapes, but the manufacturing process limits their abilities. Fabricating regions with large overhangs requires supporting structures (henceforth, supports) to prevent material collapse. While supports allow fabricating more complex models, they can be difficult to remove, waste material, and damage the model surface (ref.~\cite{Gao2015,williams1992three}). These difficulties greatly reduce the flexibility of 3D printing in automatic and agile production environments. We develop a 3D printing system that adds rotational motion into the material accumulation process to enable AM with minimal or no supports.

Various approaches try to overcome the limitation of requiring supports during the fabrication process, such as optimizing the topology of supports (e.g.,  \cite{dumas2014bridging,vanek2014clever}), searching for an optimal printing direction \cite{zhang2015perceptual}, and reducing the usage of supports by deformation \cite{Hu2015a} or model decomposition \cite{Herholz2015,muntoni2018heightblock,Hu2014}. Existing decomposition approaches often fabricate the components of a model separately, requiring a manual ``stitching'' process to obtain the final result. In other words, they cannot complete the manufacturing process in one pass, and as such, do not need to consider the constraint of collision-free fabrication. We aim to generate the 3D printing sequence for fabricating a model with one pass along different directions, which we call \textit{multi-directional 3D printing}. As shown in Fig.~\ref{figTeaser}, our approach can fabricate a model that typically requires a large area of contact supports in a support-free manner.

Our paper makes the following technical contributions:
\begin{itemize}
\item We formulate the process planning for multi-directional 3D printing as a volume decomposition problem and summarize the criteria of decomposition.

\item We propose a support-effective volume decomposition algorithm based on the beam-guided search that can be applied to general 3D models with handles and loops. 

\item We develop a region-projection based method to generate supports that are specially designed for multi-directional 3D printing to address cases where completely support-free fabrication cannot be achieved.
\end{itemize}

We developed two types of multi-directional 3D printing hardware systems: a modified off-the-shelf FDM printer with one additional rotational \textit{degree-of-freedom} (DOF) and an industrial robotic arm that simulates a tilting table, providing two rotational DOFs. Physical fabrications conducted on both systems verify the effectiveness of our method.

\section{Related Work}\label{sec-relatedWork}
Our work belongs to the interdisciplinary area of geometric computing and multi-axis fabrication, and we review the literature on model decomposition, support-oriented optimization, and multi-axis 3D printing.

\subsection{Decomposition for fabrication}
Model decomposition is a well-studied geometry processing technique that has recently seen use in 3D printing applications.

To solve the problem of printing large objects, Luo et al.\cite{luo2012chopper} design a framework to decompose large objects that exceed the working envelope of a 3D printer. They optimize the outcome of segmentation with several objective functions, such as printability, aesthetic, and structural soundness.
Vanek et al. \cite{vanek2014packmerger} propose an optimization framework that prioritizes reducing printing time and material usage by converting solids into shells and applying a packing step to merge the shells into an optimized configuration for fabrication. 
Hu et al. \cite{Hu2014} decompose a model into pyramidal parts for support-free printing. However, pyramidal decomposition is NP-hard, so they construct a weak formulation of pyramidal constraints and design an efficient algorithm to solve for the decomposition problem.
Herholz et al. \cite{Herholz2015} also try to decompose a model into parts, but instead of following a pyramidal constraint, they allow slight deformation of models to produce pieces in the shape of height-fields.
RevoMaker \cite{Gao2015UIST} can fabricate freeform models on top of an existing electronic component in a cubic shape. Again, the models need to be decomposed into the shape of height-fields. 
Yao et al. \cite{yao2015level} develop a level-set method to deal with the problems of partitioning and packing. First, mesh segmentation constructs an initial volume decomposition. Then, their results undergo alternating optimization via an iterative variational optimization method, where partitioning and packing energies are defined in volumetric space.
Chen et al. \cite{Chen2015dapper} also decompose an input model into a small number of parts that can be efficiently packed for 3D printing. They use an algorithm that explores the decomposition and packing space with a prioritized and bounded beam search guided by local and global objectives.

Staircase artifacts generated by layer-based printing are considered a major type of defect in 3D printed models. To solve this problem, \cite{wang2016improved} creates a method for subdividing the shape into parts that can be built in different directions. After printing all the individual parts, the 3D printed model is manually assembled, which improves the visual quality by reducing the staircase artifacts.
Song et al. \cite{song2016cofifab} propose an approach for fabricating large-scale models by combining 3D printing and laser cutting in a coarse-to-fine fabrication process.
Wei et al. \cite{wei18supportfree} present a skeleton-based algorithm for partitioning a 3D shell model into a small number of support-free parts, each of which has a specific printing direction that leads to support-free fabrication. The method also minimizes seams and cracks by integrating the length of the cuts into the optimization formulation.
Muntoni et al. \cite{muntoni2018heightblock} recently proposed a decomposition algorithm for processing general 3D objects into a small set of non-overlapping height-field blocks. The directions of the height-fields are constrained to the principal axes to solve the overlapping problem. These blocks can be fabricated by moulding or 3D printing.

None of the discussed decomposition approaches considers the collision-free constraint and sequence of manufacturing.
Therefore, they cannot be directly applied to our multi-directional 3D printing system.

\begin{figure*}[t]
\centering
\includegraphics[width=\linewidth]{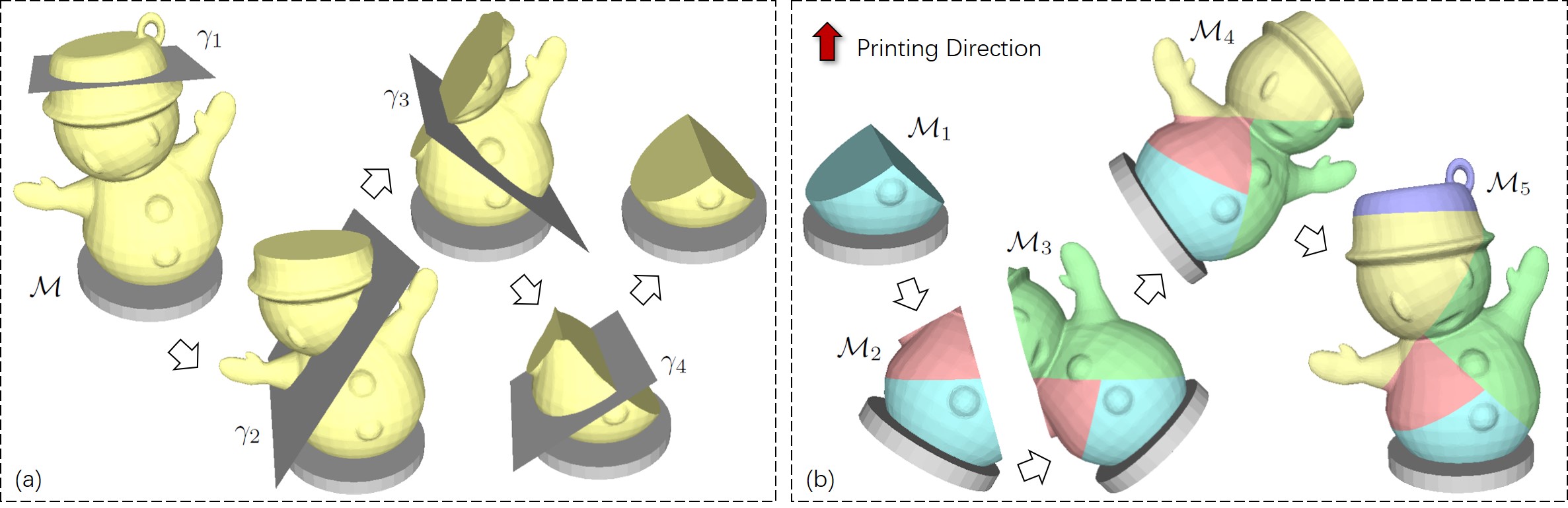}
\caption{An illustration for our algorithm: (a) progressively determined results of planar clipping for generating the optimized decomposition, and (b) the inverse order of clipping planes that result in a sequence of regions for fabrication. The printing direction of each region is the normal of its base plane. Note that the orientation of a printing head is fixed during the procedure of physical fabrication, and the parts under fabrication are reoriented to realize the multi-directional 3D printing.
} 
\label{figAlgOverview}
\end{figure*}

\subsection{Support-oriented optimization}
Though AM claims to have the ability to fabricate models with complex shapes, the need for supports reduces the flexibility of production. Many prior approaches aim at optimizing either the efficiency of production or the appearance of the model. 
Several previous works use the volume of supports as an optimization objective for generating effective supports.
Vanek et al. \cite{vanek2014clever} propose an algorithm for generating hierarchical support structures. MeshMixer \cite{Branching} also provides a well-designed hierarchical pattern to generate efficient support structures. Dumas et al. \cite{dumas2014bridging} introduce a bridge-like support structure generation algorithm. Bridges are stronger and more stable than hierarchical structures while still maintaining manufacturing efficiency. However, with the development of fast 3D printing technology, the contact area of the supports becomes more critical than the volume, which is because a larger contact area requires more effort when manually removing the supports, and the final model retains more surface artifacts.

Another thread of research focuses on changing the 3D printing orientation to reduce the contact area of supports. Hu et al. \cite{Hu2015a} design an orientation-driven shape deformation framework to adaptively adjust the orientation of regions with large overhangs. Zhang et al. \cite{zhang2015perceptual} propose a double-layered perceptual neural network, \emph{DL-ELM}, to rank a list of possible printing directions, with the expectation that the best printing direction prevents critical visual features from being damaged by additional supports. %
Similar to these works, we introduce rotations into 3D printing to reduce the number of required supports by minimizing their contact area.

\subsection{Multi-axis 3D printing}
Layer-based approaches heavily restrict the flexibility and efficiency of 3D printing. We review methods for adding more DOFs into the 3D printing process.

Keating and Oxman \cite{Keating2013} present a manufacturing platform using 6-DOF provided by a robotic arm to fabricate models in both additive and subtractive manners. 
Pan et al.~\cite{Pan2014} propose a 5-axis motion system similar to 5-axis CNC machines to accumulate materials. A 6-DOF parallel kinematic Stewart platform is presented in the work of Song et al.~\cite{Song2015} for multi-directional AM. These systems only fabricate small models with simple shapes.
Peng et al. \cite{peng2016fly} propose an \emph{On-the-Fly Print} system to enable fast, interactive fabrication by adding a 2-DOF rotational platform to an off-the-shelf Delta 3D printer. This system can fabricate both solid and wireframe models.
Using the \emph{On-the-Fly Print} system, Wu et al. \cite{wu2016printing} propose an algorithm to generate a collision-free printing order for edges in wireframe printing.
Huang et al. \cite{huang2016framefab} build a robotic arm 3D printing system based on a 6-DOF KUKA robotic arm and a customized extrusion head. They also propose a divide-and-conquer algorithm to search for a possible fabrication sequence that is both structurally stable and collision-free.
Dai et al. \cite{dai2018support} recently developed a support-free volume printing system equipped with a 6-DOF robotic arm.
Shembekar et al.~\cite{shembekar2018trajectory} present a method for conducting conformal 3D printing of freeform surfaces by collision-free trajectories, and this method has been validated on a 6-DOF robotic arm. 
These approaches deposit materials along 3D tool paths and require relatively expensive devices and control systems to move all DOFs together during the fabrication process. In contrast, our approach decouples the motion for changing orientation from the motion for 3D printing. As a result, the decomposition generated by our algorithm can be used to supervise the fabrication of general models on a device with meager cost (e.g., the system introduced in Section~\ref{subsecCartesianSystem}).

The decomposition work presented in \cite{xu18supportfree} for $3+2$-axis additive manufacturing relates closely to our approach. Their work employs a flooding algorithm to segment a given mesh surface into different regions, which can then be fabricated along with different directions without supports. However, this approach is limited in that it only works for tree-like models with simple topologies. We propose a more general approach that can process non-tree-like models as well as models with handles and loops. Discussion and comparison with \cite{xu18supportfree} can be found in Section~\ref{subsecResDiscussion}.

\section{Methodology}\label{secMethodology}

\subsection{Problem Statement}
Given a model $\mathcal{M}$ fabricated layer-by-layer on a base plane $\pi$ with a printing direction $\mathbf{d}_{\pi}$, identify whether a face $f$ with normal $\mathbf{n}_f$ is self-supported by

\begin{equation}
e(f, \pi)=
\begin{cases}
1 & \mathbf{n}_f \cdot \mathbf{d}_{\pi} + \sin{( \alpha_{max} ) < 0}, \\
0 & \text{otherwise}.
\end{cases}
\end{equation}
where $\alpha_{max}$ is the maximal self-supporting angle (ref.\cite{Hu2015a}). Face $f$ is a \textit{risky} face w.r.t $\pi$ when $e(f,\pi) = 1$, otherwise it is a \textit{safe} face. Note that $\mathbf{d}_{\pi}$ is the normal of $\pi$. Clearly, the need for supports relates strongly with the printing direction, providing the opportunity for reducing or eliminating supports by changing printing direction during manufacturing.

To supervise the operating multi-directional 3D printer, we need to generate a decomposition of $\mathcal{M}$ where:
\begin{itemize}
\item $\mathcal{M}$ has $N$ components such that
\begin{equation}
\mathcal{M} = \mathcal{M}_1 \cup \mathcal{M}_2 \cup \cdots \cup \mathcal{M}_N = \cup_{i=1}^{N} \mathcal{M}_i
\end{equation}
with $\cup$ denoting the \textit{union} operator;

\item $\{ \mathcal{M}_{i=1,\cdots,N} \}$ is an ordered sequence that can be collision-freely fabricated with \begin{equation}
\pi_{i+1} = \mathcal{M}_{i+1} \cap \left( \cup_{j=1}^{i} \mathcal{M}_j \right) 
\end{equation}
being the base plane of $\mathcal{M}_{i+1}$ -- here $\cap$ denotes the \textit{intersection} operator;

\item $\pi_1$ is the working platform of a 3D printer;

\item All faces on a sub-region $\mathcal{M}_i$ are \textit{safe} according to $\mathbf{d}_{\pi_i}$ determined by $\pi_i$.
\end{itemize}

We solve the weak-form problem by reducing the area of risky faces on each component $\mathcal{M}_i$. That is, minimize
\begin{equation}\label{eqGlobalObjective}
J_{G} = \sum_{i} \sum_{f \in \mathcal{M}_i} e(f, \pi_i) A(f)
\end{equation}
where $A(f)$ is the area of face $f$. While minimizing the objective function (Eq.(\ref{eqGlobalObjective})), we need to ensure the fabricating each component is collision-free. 
For regions where faces are not entirely safe, we generate supports specially designed for multi-directional 3D printing.

\subsection{Our Approach}\label{subsecMethodOurApproach}
Multi-directional 3D printing a given model $\mathcal{M}$ requires determining an ordered sequence of clipping planes $\gamma_k$ ($k=1,\ldots,N-1$) that decomposes $\mathcal{M}$ into $N$ components (see Fig.\ref{figAlgOverview}(a)).
We define the half-space of a clipping plane $\gamma_k$ containing the 3D printing platform $\mathcal{P}$ as `below', denoted by $\Gamma_k^-$) and the half-space `above' $\gamma_k$ is $\Gamma_k^+$. The clipping operation gives the remained model by
\begin{equation}\label{eqClipping}
\bar{\mathcal{M}}_k =  \bar{\mathcal{M}}_{k-1} \setminus \Gamma_k^+ 
\end{equation}
with $\bar{\mathcal{M}}_0 = \mathcal{M}$ and `$\setminus$' denotes the subtraction operator on solids. When every clipped sub-region in $\Gamma_k^+$ satisfies the criteria of manufacturability (see Section \ref{subsecCriteria} below), the inverse order of clipping gives the sequence of region printing for multi-directional 3D printing. Specifically, we have 
\begin{equation}\label{eqInverseClipping}
\mathcal{M}_i=\bar{\mathcal{M}}_{(N-i)} \cap \Gamma^+_{(N-i+1)}, \quad \pi_i = \gamma_{(N-i+1)} %
\end{equation}
with $i=1,\cdots,N$. The printing direction of a sub-region $\mathcal{M}_i$ is given by the normal vector of $\gamma_k$ pointing from $\Gamma_k^-$ into $\Gamma_k^+$ with $k = N-i+1$. See Fig.\ref{figAlgOverview}(b) for an illustration of using the inverse order of clipping to obtain the sequence for multi-directional 3D printing. Half-spaces defined by sequentially applying all $N$ clipping operations subdivide the $\Re^3$ space into $N+1$ convex sub-space. The first $k$ clipping operations generate the $\mathcal{M}_i$ component (Eq.(\ref{eqInverseClipping})) in a sub-space as in
\begin{equation}\label{eqSpaceOfComponent}
\Omega_k = \Gamma^-_1 \cap \Gamma^-_2 \cap \cdots \cap \Gamma^-_{k-1} \cap \Gamma^+_k = \left( \cap_{j=1}^{k-1} \Gamma^-_j \right) \cap \Gamma^+_k
\end{equation}
with $\Omega_1=\Gamma^+_1$. When needed, the supporting structure for the component $\mathcal{M}_i$ will be generated in $\Omega_k$ ($k = N-i+1$) and progressively projected into the rest sub-space $\Omega_{j}$ ($j>k$) until it can be merged with other supports or meets the printing platform $\mathcal{P}$ (see Section \ref{secSupport} for details).

Candidates of clipping planes can be generated by 
\begin{enumerate}
\item uniformly sampling the Gaussian sphere to obtain 250 normals and
\item applying a uniform shifting along each sampled normal vector with an offset of $1mm$.
\end{enumerate}
For all examples shown in this paper, this sampling strategy generated around $15\mathrm{k} \sim 20\mathrm{k}$ candidate clipping planes. We develop a beam-guided search scheme to select an optimized order of clipping, which can significantly improve the local-optimum results obtained from a greedy scheme. Details will be presented in Section \ref{secSearchScheme}.

The planar clipping methodology employed in this work can process general models with a high-genus number, addressing the drawbacks in \cite{wu2017RoboFDM}, \cite{wei18supportfree}, and \cite{xu18supportfree}, which could only process models with skeletal-tree structures.
Additionally, the algorithm can be easily tailor-made to support a hardware system with only one rotational axis (e.g., the system shown in Fig.\ref{figHardwareSystem}(a)). This is realized by generating samples $\mathbf{n}_k$ on a circle of the Gaussian sphere satisfying $\mathbf{n}_k \cdot \mathbf{r} = 0$ with $\mathbf{r}$ being the axis of rotation. Moreover, we provide a support generation solution to enable the fabrication of all models on a multi-directional 3D printing system.

\subsection{Criteria for Decomposition}\label{subsecCriteria}
We now define the criteria for finding an optimal sub-region $\mathcal{M}_{i}$ according to a clipping plane $\gamma_k$ ($k=N-i+1$) for multi-directional 3D printing. Here $\pi_i$ denotes the corresponding base plane of $\gamma_k$.

\vspace{5pt} \noindent \textbf{Criterion I:} $\forall f \in \mathcal{M}_i$, $e(f,\pi_i)=0$ -- i.e. all faces on $\mathcal{M}_i$ are self-supported. \vspace{5pt}

\noindent Minimization of the objective function (Eq.(\ref{eqGlobalObjective})) imposes this criterion, ensuring the manufacturability of the region above $\gamma_k$.

\vspace{5pt} \noindent \textbf{Criterion II:} The model $\bar{\mathcal{M}}_k$ obtained from the clipping by $\gamma_k$ must be connected to the printing platform $\mathcal{P}$.
\vspace{5pt}

\noindent This criterion prevents unmanufacturable configurations for the region below $\gamma_k$ by avoiding the generation of ``floating'' regions, which require supports when printed.

The next criterion avoids collisions between the printer head and the platform.

\vspace{5pt} \noindent \textbf{Criterion III:} The printing platform $\mathcal{P}$ and the clipping plane $\gamma_k$ satisfy $\Gamma_k^+ \cap \mathcal{P} = \emptyset$ (i.e., $\mathcal{P}$ is below $\gamma_k$). \vspace{5pt}

\noindent Note that we do not explicitly prevent collision between the printer head and already fabricated regions as the clipping routine that generates sub-regions from $\mathcal{M}$ already guarantees this. 
All regions $\mathcal{M}_{j\,(j<i)}$ are below the base plane $\pi_i$ (i.e. the clipping plane $\gamma_{(N-i+1)}$) because the sequence of 3D printing is the inverse order of clipping.

\begin{figure*}[t]
\centering
\includegraphics[width=\linewidth]{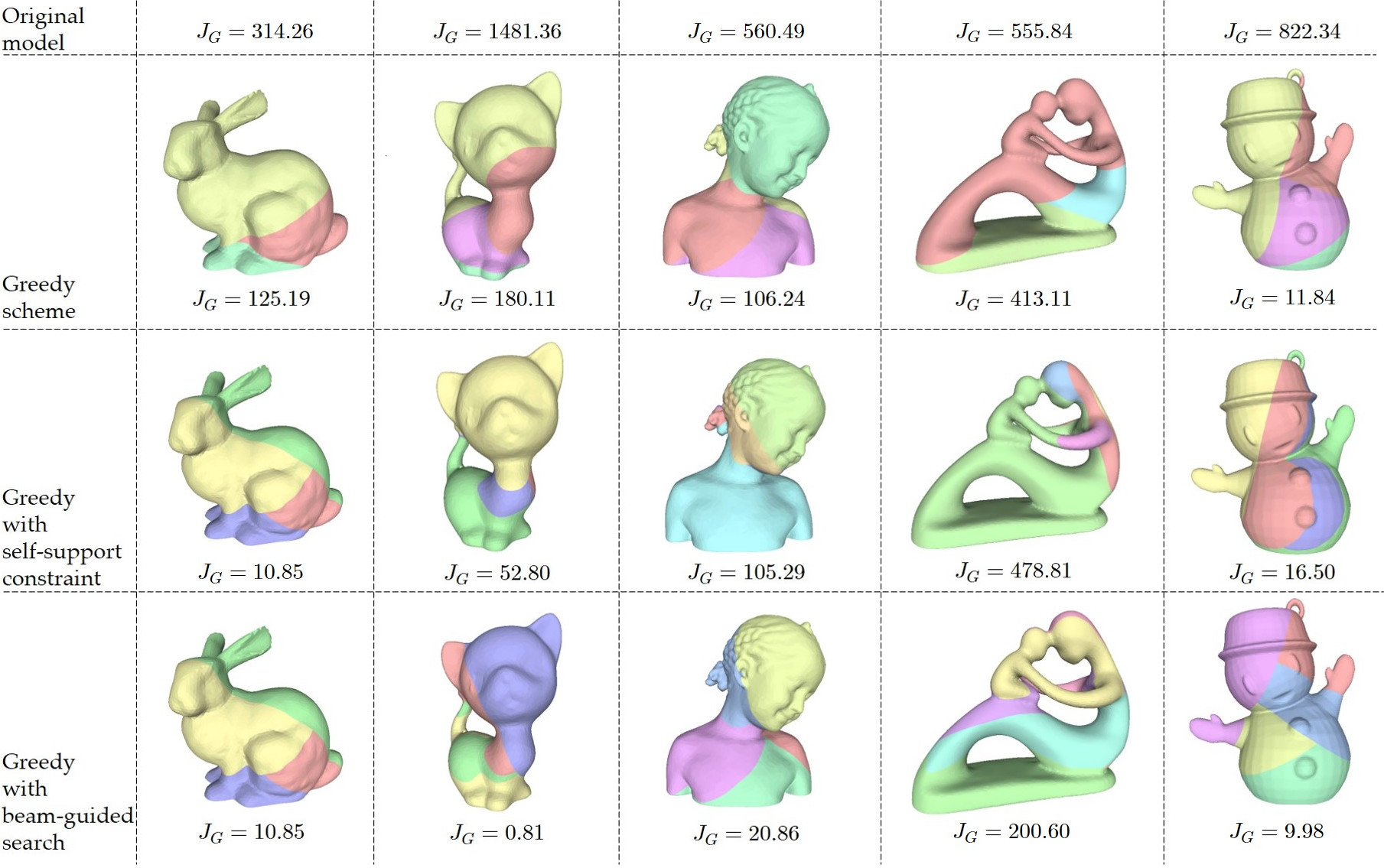}
\caption{A comparison of decomposition results obtained from three schemes introduced in this paper. Beam-guided search always determines the ``best'' decomposition (i.e., the one with minimized $J_G$ (Eq.(\ref{eqGlobalObjective})).}
\label{figSchemeComparison}
\end{figure*}

In practice, we cannot always find a decomposition satisfying Criterion I for all components. For such a scenario, a weak form for support-free is adopted to aim at generating a smaller area of overhang during fabrication.

When changing from one printing direction to another printing direction, the following drawbacks are introduced:
\begin{itemize}
\item The visual artifact of a curve is formed at the interface of two neighboring regions;

\item It takes extra time for the machine to move from one orientation to the other -- the printing process is slower.
\end{itemize}
Therefore, we generally prefer a solution with fewer components, which can be achieved by considering the following criterion of clipping.

\vspace{5pt} \noindent \textbf{Criterion IV:} We prefer a large solid volume for the region above a clipping plane.\vspace{5pt}

In summary, the volume decomposition of $\mathcal{M}$ is considered \textit{optimized} when it satisfies all the above criteria. Two schemes developed in the next section compute the optimal decomposition.

\section{Schemes of Optimization}\label{secSearchScheme}
We introduce greedy and beam-guided search schemes for determining the sequence of clipping planes and thereby obtaining the corresponding volume decomposition. The beam-guided search scheme avoids converging at local optima.

\subsection{Greedy Scheme}\label{subsecGreedySearch}
Consider a clipping plane $\gamma$ that decomposes the current model $\bar{\mathcal{M}}_c$ into the upper and lower parts, $\bar{\mathcal{M}}_c^+$ and $\bar{\mathcal{M}}_c^-$. A greedy scheme searches for a $\gamma$ that most significantly reduces the current value of the global objective function (i.e., $J_G$ in Eq.(\ref{eqGlobalObjective})). To implement this strategy, we define a local objective function as a weak form of Criterion I, which evaluates the descent of risky areas, by:
\begin{eqnarray}
J_{L} &=& \sum_{f \in \bar{\mathcal{M}}_c} e(f, \pi(\mathcal{P}))A(f) -  \\
    &&  \left(  \sum_{f \in \bar{\mathcal{M}}_c^-} e(f, \pi(\mathcal{P}))A(f) + \sum_{f \in \bar{\mathcal{M}}_c^+} e(f, \pi(\gamma))A(f) \right)  \nonumber \\
    & =&  \sum_{f \in \bar{\mathcal{M}}_c^+} (e(f, \pi(\mathcal{P})) - e(f, \pi(\gamma)) )A(f)  \nonumber
\end{eqnarray}
where $\mathcal{P}$ is the platform of a 3D printer.

Among all the candidate clipping planes, the greedy scheme always selects the plane with maximum $J_{L}$:
\begin{equation}\label{eqUnconstrainedGreedy}
\gamma = \arg \max J_{L} (\gamma).
\end{equation}
However, such a selection does not guarantee the region $\bar{\mathcal{M}}_c^+$ above the selected clipping plane is completely support-free. 
To address this, %
we propose a constrained greedy scheme. The clipping plane is selected by
\begin{equation}\label{eqConstrainedGreedy}
\gamma = \arg \max J_{L} (\gamma) \quad s.t. \;\; R(\mathcal{M}_c^+,\gamma) = 0
\end{equation}
with 
\begin{equation}\label{eqRiskyRegion}
    R(\mathcal{M}_c^+,\gamma)= \sum_{f \in \bar{\mathcal{M}}_c^+} e(f, \pi(\gamma))A(f)
\end{equation}
giving the total area of risky faces on $\mathcal{M}_c^+$. %

In practice, we first select candidates among the clipping planes that let $\mathcal{M}_c^+$ be completely self-supported. If there is no such a clipping plane, we solve a degenerated problem (Eq.(\ref{eqUnconstrainedGreedy})). After adding this preference for support-free regions, the objective function $J_G$ can be further minimized on most models (as shown in Fig.\ref{figSchemeComparison}). However, counterexamples can also be found where the additional constraint increases $J_G$ -- e.g., the snowman. This is mainly because a greedy scheme can easily fall into a local optimum. A better search scheme needs to be developed by considering all the criteria discussed in Section \ref{subsecCriteria}.

\subsection{Beam-Guided Search Scheme}\label{subsecBeamGuidedSearch}
As aforementioned, in many cases, it is not guaranteed to find a support-free decomposition for every sub-region -- i.e., Criterion I is not satisfied for some regions. To provide a general solution, we reformulate this criterion into a weak form as a local objective function. Specifically, we search for a clipping plane $\gamma_k$ that leads to
\begin{equation}\label{eqObjMinArea}
\min R(\mathcal{M}_{N-k+1},\gamma_k)
\end{equation}
with $R(\cdot,\cdot)$ evaluating the total area of risky faces as defined in Eq.(\ref{eqRiskyRegion}). 
Criterion II and III are imposed by excluding those unsatisfactory clipping planes from the set of candidates. Similarly, to avoid generating too many small fragments when decomposing an input model $\mathcal{M}$, clipping operations that lead to a sub-region with volume less than $V(\mathcal{M}) / w$ are prevented. Here, $w$ is a user-specified parameter to control the maximal number of components (i.e., $w$ from $10$ to $12$ is used in our tests). %

Beam search \cite{Lowerre1976} is an efficient search technique that has been widely used to improve the results of best-first greedy search. A breadth-first strategy is employed to build a search tree that explores the search space by expanding the set of most promising nodes instead of only the best node at each level. It has been successfully used in a variety of areas, especially in geometric configuration search tasks for 3D printing (e.g., \cite{luo2012chopper,Chen2015dapper}). Our approach introduces a progressive relaxation routine to conduct the breadth-first search.

The most challenging part in solving our volume decomposition is integrating the restrictive Criterion I (and its weak form) as an objective function presented in Eq.(\ref{eqObjMinArea}). This important step ensures that the beam search is broad enough to include both the local optimum and configurations that may lead to a global optimum. In contrast with the traditional usage of a beam search algorithm that keeps the $b$ most promising results, our beam-guided search algorithm starts from an empty beam with the most restrictive requirement of  $R(\mathcal{M}_{N-k+1},\gamma_k) < \delta$, where $\delta$ is a tiny number (e.g., $\delta=0.1$ is used in all our tests). 
Candidate clipping planes that satisfy this requirement and remove larger areas of risky faces have higher priority when filling the $b$ beams. If there are still empty beams, we relax $\delta$ by letting $\delta=5\delta$ until all $b$ beams are filled. Details of our beam-guided search algorithm are presented below.

\begin{algorithm}[t]
\caption{\texttt{Beam-Guided Search}}\label{algBeamGuidedSearch}

\LinesNumbered

\KwIn{Input mesh $\mathcal{M}$.}
\KwOut{An optimized set of decomposition as $\{\mathcal{M}_i\}$.}

Build a set of uniformly sampled candidate planes $\Pi$ (Section \ref{subsecMethodOurApproach}) that $\forall \gamma \in \Pi, \, \gamma \cap \mathcal{P}=\emptyset$ (Criterion III);

Initialize $b$ empty beams as $\forall j, \, \mathcal{B}^j=(null,null)$; 

$\mathcal{B}^1=(\mathcal{M},null)$;

\Repeat{$\forall j$, $\mathcal{M}(\mathcal{B}^j) = null$}{

    \tcc{Preparing clipping candidates}
    Initialize a maximum heap of clipping as $\mathcal{G}=\emptyset$;

    \ForAll{$\mathcal{B}^j$}{
        \If{$\mathcal{M}(\mathcal{B}^j) \neq null$}{
            \ForAll{$\gamma \in \Pi$}{
                The clipping plane $\gamma$ decomposes $\mathcal{M}(\mathcal{B}^j)$ into $\mathcal{M}^+$ and $\mathcal{M}^-$;
                
                \If{$(V(\mathcal{M}^+) \geq \frac{1}{w}V(\mathcal{M}))$ %
                    \texttt{AND}
                    $(\mathcal{M}^-$ is connected to $\mathcal{P})$}{   %
                        Evaluate $J_G$ according to $\mathcal{L}(\mathcal{B}^j)$;

                        Insert the tuple $(\mathcal{M}^-,\mathcal{M}^+,\mathcal{L}(\mathcal{B}^j),\gamma)$ into $\mathcal{G}$ with $J_G$ as the key;  %
                }
            } 
        }
    }
    
    \tcc{Progressive relaxation}
    $\delta = 0.1$ and $k=0$;
    
    \While{$k<b$}{
        Initialize a maximum heap $\bar{\mathcal{G}}$ as buffer;
        
        \While{$\mathcal{G} \neq \emptyset$}{
            Pop $(\mathcal{M}^-,\mathcal{M}^+,\mathcal{L},\gamma)$ from the top of $\mathcal{G}$;
    
            \eIf{$R(\mathcal{M}^+,\gamma)<\delta$}{
                Add $\gamma$ at the tail of $\mathcal{L}$;
            
                Let $\mathcal{B}^{k+1} = (\mathcal{M}^-,\mathcal{L})$ and $k=k+1$;
            }
            {
                Evaluate $J_G$ according to $\mathcal{L}$;

                Insert the tuple $(\mathcal{M}^-,\mathcal{M}^+,\mathcal{L},\gamma)$ into $\bar{\mathcal{G}}$ with $J_G$ as the key;
            }
        }
        Let $\mathcal{G}=\bar{\mathcal{G}}$ and $\delta = 5 \delta$;
    }
    
    \tcc{Checking terminal condition}
    \ForAll{$\mathcal{B}^j$}{
        \If{$(V(\mathcal{M}(\mathcal{B}^j))<\frac{1}{w} V(\mathcal{M}))$ 
            \texttt{OR} $(R(\mathcal{M}(\mathcal{B}^j),\pi(\mathcal{P}))=0)$}{

            Evaluate $J_G$ according to $\mathcal{L}(\mathcal{B}^j)$;
            
            $\mathcal{B}^j=(null, \mathcal{L}(\mathcal{B}^j))$;
        }
    }
}

\textbf{return} the decomposition that gives the minimal $J_G$;

\end{algorithm}

The algorithm starts from $b$ empty beams $\mathcal{B}^j$ ($j=1,\ldots,b$), where each beam $\mathcal{B}^j=(\mathcal{M}(\mathcal{B}^j),\mathcal{L}(\mathcal{B}^j))$ contains a remaining model $\mathcal{M}(\mathcal{B}^j)$ and an ordered list of clipping planes, $\mathcal{L}(\mathcal{B}^j)$, that forms $\mathcal{M}(\mathcal{B}^j)$. In our implementation, only the last element of a list needs to be stored in $\mathcal{B}^j$, as the rest of the prior elements in the list can be traced through a backward link. By using the progressive relaxation routine, each beam can be extended by adding a new clipping plane into its list and obtaining an updated remaining model. In the next round, apply progressive relaxation to all valid clipping results for all $b$ remaining models, prioritizing the removal of more risky faces. Repeat the extension process of beam $\mathcal{B}^j$ until any of the following terminal conditions is satisfied on the remaining model $\mathcal{M}(\mathcal{B}^j)$
\begin{itemize}
\item \textit{Small Volume}: The volume of the remaining model $\mathcal{M}(\mathcal{B}^j)$ has small volume -- i.e., $V(\mathcal{M}(\mathcal{B}^j))<\frac{1}{w} V(\mathcal{M})$;

\item \textit{Self-Supported}: The remaining model $\mathcal{M}(\mathcal{B}^j)$ is completely self-supported as $R(\mathcal{M}(\mathcal{B}^j),\pi(\mathcal{P}))=0$.
\end{itemize}
The search also terminates when no beam can be further extended.

Each beam corresponds to a list of clipping planes that gives a decomposition model $\mathcal{M}$. The decomposition that leads to the minimal value of $J_G$ (Eq.(\ref{eqGlobalObjective})) is considered the optimized solution for our multi-directional 3D printing. Pseudo-code of our beam-guided search algorithm is given in Algorithm \ref{algBeamGuidedSearch}. We use the method of discarding results close to already selected ones, which is proposed in~\cite{luo2012chopper}, to avoid filling the beam by similar results. Example results and a comparison against two other greedy (constrained and unconstrained) schemes can be found in Fig.\ref{figSchemeComparison}. The beam-guided search gives the best decomposition on all models.

\section{Support Generation}\label{secSupport}
After relaxing the hard-constraint of creating a support-free model decomposition into minimizing the area of risky faces, $J_G$, the scheme for generating supports is considerately important for models that still have risky faces after decomposition. To tackle this problem, we propose a new structure called \textit{projected support} that ensures the fabrication of remaining overhanging regions through collision-free multi-directional 3D printing.

For a decomposition that results in a sequence of sub-regions $\mathcal{S}=\{ \mathcal{M}_i \}$ with base plane $\pi_i$ corresponding to each sub-region $\mathcal{M}_i$, the printing direction $\mathbf{d}_i$ is the normal vector of $\pi_i$. 
Thus, the overhanging region on $\mathcal{M}_i$ with respect to the printing direction $\mathbf{d}_i$ can be detected and supporting structures added along the direction $(-\mathbf{d}_i)$. Here, we select the tree-like support \cite{vanek2014clever} that merges the supporting structures for different overhanging regions when they are near each other. In fact, our progressive projection algorithm is general and can be applied to different patterns of supports, such as the bridge-like support \cite{dumas2014bridging} or other denser supports \cite{huang2014Image}.

\begin{figure}
	\centering
    \subfigure[Tree-like supports]{ 
		\includegraphics[width=0.4\linewidth]{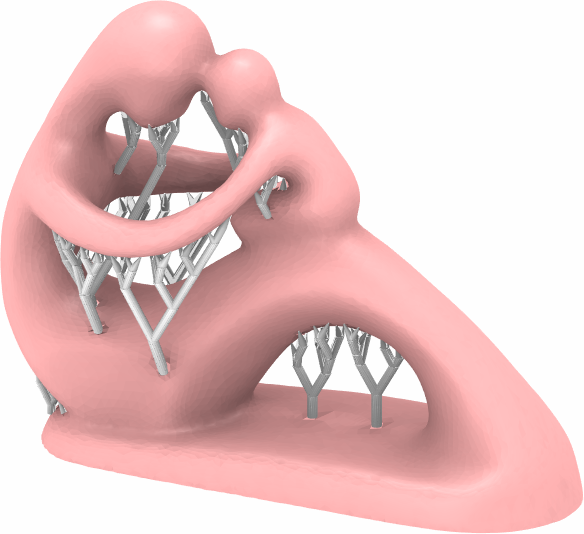}
		}
	\subfigure[Projected supports]{ 	
		\includegraphics[width=0.4\linewidth]{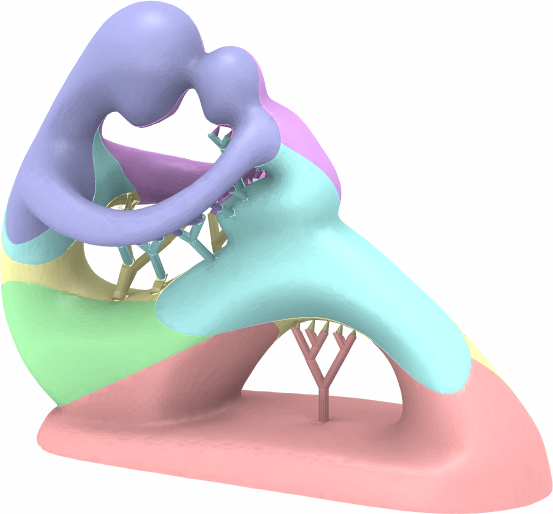}
     }
\caption{The sparse tree-like supporting structures for 3D printing along with a fixed direction (a) and the progressively projected supports generated by our algorithm for multi-directional 3D printing (b). Note that, fewer supports are needed for the multi-directional printing. To avoid the issues of stability raised by gravitational torques, we incorporate dynamic struts defined in Eq.(\ref{eqSupportBalance}) to generate projected support structures in our system.}\label{figSupportComp}
\end{figure}

Unlike the existing algorithms that generate supports by projecting along a fixed printing direction, in our case, the projection should be conducted along with different directions in different regions. Without loss of generality, the $i$-th component $\mathcal{M}_i$ falls in a sub-space $\Omega_k$ formed by the first $k$ clipping planes -- as shown in Eq.(\ref{eqSpaceOfComponent}) with $k=N-i+1$. 
The support for the overhang on $\mathcal{M}_i$ is generated along the inverse printing direction $(-\mathbf{d}_i)$ and projected onto the base plane (i.e., the $k$-th clipping plane, $\gamma_k$). Next, the structure is projected into a new sub-space $\Omega_{k+1}$ and along a new direction $(-\mathbf{d}_{i-1})$. The projection repeats until this structure can be merged with other structures or meets the printer's platform $\mathcal{P}$. Fig.\ref{figSupportComp}(b) shows supports generated by our progressive projection algorithm displayed in different colors when they are in different sub-spaces $\{ \Omega_k \}$. The pseudo-code of our progressive projection algorithm is given in Algorithm \ref{algProgressProjection}.

After applying the volume decomposition algorithm, we use the uniformly sampling strategy to detect all overhang types -- including point-overhang, edge-overhang and face-overhang. Sampling interval $R$ is a parameter selected by the diameter of the deposition nozzle and struts. We use $R=3$mm for printing with a $0.8$mm diameter nozzle and $R=2$mm for printing with a $0.4$mm diameter nozzle. 
For each sub-mesh $\mathcal{M}_s$ and its associated sub-space $\Omega_s$, an ideal configuration of support structures $\mathcal{C}_s$ should satisfy $\mathcal{C}_s \subset \Omega_s$ with a minimal number of points that contact the input model. 
We use the method proposed in \cite{vanek2014clever} to ensure that the newly generated nodes of the tree are inside $\Omega_s$ during the merging procedure. This guarantees that the corresponding connected structures are inside the convex space $\Omega_s$. We set the maximal self-supporting angle for sparse tree-like supports to $30^\circ$ and adopt a heuristic greedy-based method~\cite{vanek2014clever} to progressively merge pairs of supporting structures when they are close to each other.

Moreover, considering the stability issue raised by gravitational torque when printing along with different directions, we propose the following function for selecting the diameter $R_p$ of a projected supporting strut $\mathcal{C}_p \subset \Omega_k$.
\begin{equation}\label{eqSupportBalance}
    R_p = (1 + \lambda  \| \sum_i V_i (\mathbf{c}_i \times \mathbf{g}) \| ) R/4
\end{equation}
where $\| \sum_i V_i (\mathbf{c}_i \times \mathbf{g}) \|$ is the torque on top of $\mathcal{C}_p$, and  $\mathbf{c}_i$ and $V_i$ are the centroid and the volume of the supporting strut connected to $\mathcal{C}_p$. $\lambda$ is a user-defined parameter to determine the diameter of projected supports, and we empirically set it as $10^{-6}$.

\begin{algorithm}[t]
\caption{\texttt{Progressive Projection}}\label{algProgressProjection}

\LinesNumbered

\KwIn{Components of $\mathcal{M}$ in a sequence $\mathcal{S}$.}
\KwOut{Support structures $\mathcal{T}$}

Initialize an empty set $\mathcal{T}=\emptyset$ for support-structures;

\For{$i=N,\ldots,1$}{
    
    \If{$R(\mathcal{M}_i,\pi_i)>0$}{ 
        Generate support $\mathcal{C}_i$ for $\mathcal{M}_i$ inside $\Omega_{N-i+1}$;
    
        Merge $\mathcal{C}_i$ into $\mathcal{T}$;
    }
    
    Extend $\mathcal{T}$ along the direction of $(-\mathbf{d}_i)$ until meeting the base plane $\pi_i$ or the component $\mathcal{M}_i$;
}

\textbf{return} $\mathcal{T}$;
\end{algorithm}

\section{Experimental Results}\label{secResult}
We have implemented the proposed search algorithms in C++ and Python programs and tested them on a PC with two Intel E5-2698 v3 CPUs and 128GB RAM. To prove the effectiveness of our algorithm, we use a conservative choice of the maximal self-supporting angle as $\alpha_{\max}=45^\circ$. In practice, this parameter highly depends on a 3D printer's capability, and up to $70^\circ$ can be achieved by advanced 3D printers, such as the one used in \cite{wei18supportfree}. The slicing software for conventional FDM, Ultimaker Cura~\cite{ultimakerCura}, is used to create planar slices and tool-paths according to the printing directions determined in our algorithm. The generated g-code for fabrication is sent to the motion-control module of the hardware. Two different hardware platforms have been used to verify the effectiveness of decomposition, with different configurations for navigating the motions of material extrusions. One is a Cartesian-space-based system with 4-DOF in motion, and the other hardware platform is a joint-space-based system~\cite{wu2017RoboFDM} that is built on an industrial robotic arm equipped with a fixed FDM extruder. 

\begin{figure}[t]
\includegraphics[width=\linewidth]{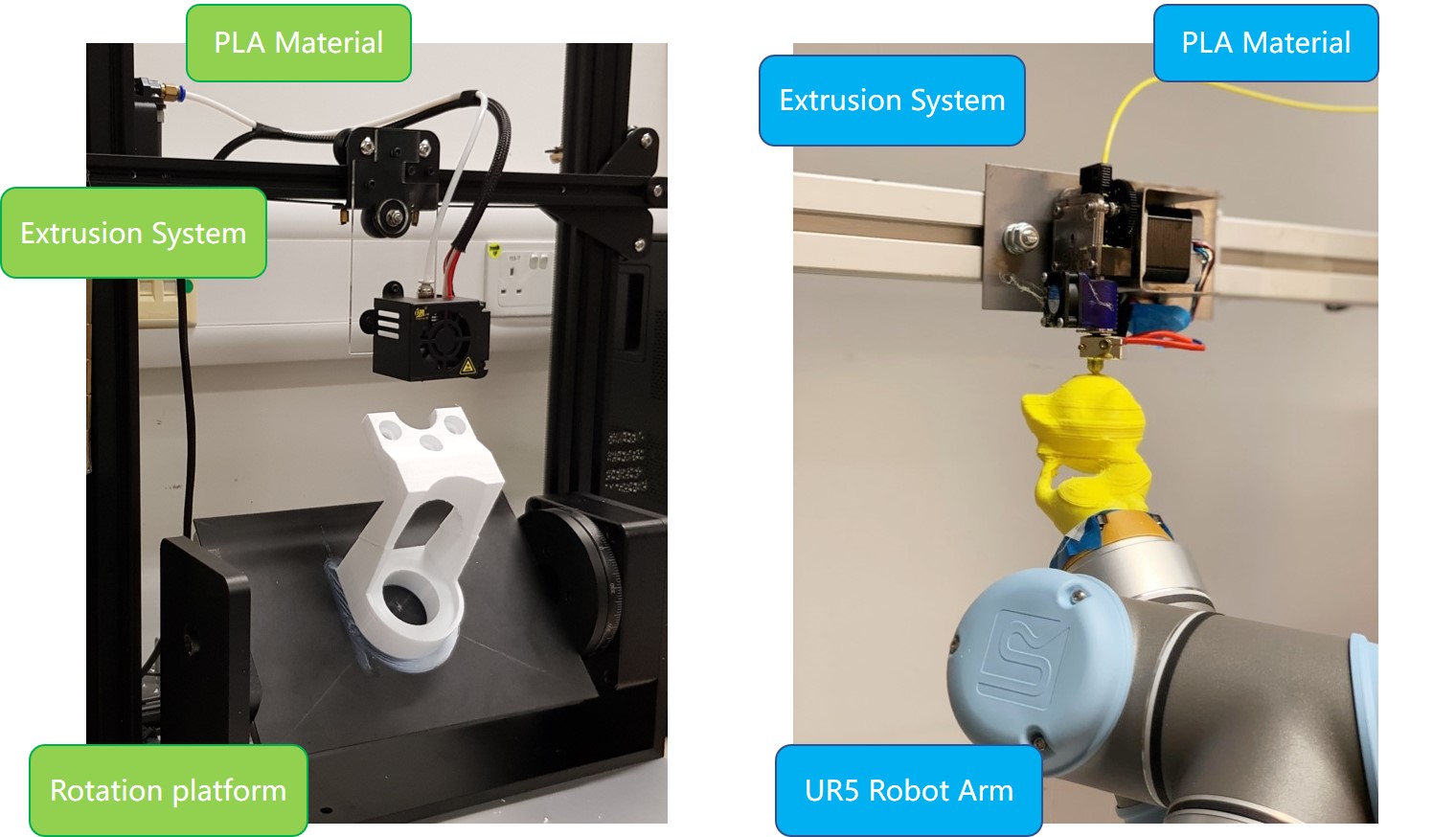}
\caption{Two different hardware setups for multi-directional printing have been built to verify the effectiveness of our volume decomposition approach. (Left) A Cartesian-space-based 4DOF printer modified from an off-the-shelf FDM printer with an additional degree-of-freedom to provide the capability of rotation. (Right) A joint-space-based 5DOF system consisting of an industrial 6DOF robotic arm and a fixed FDM extruder. %
} 
\label{figHardwareSystem}
\end{figure}
\begin{figure}[t]
\centering
\includegraphics[width=0.7\linewidth]{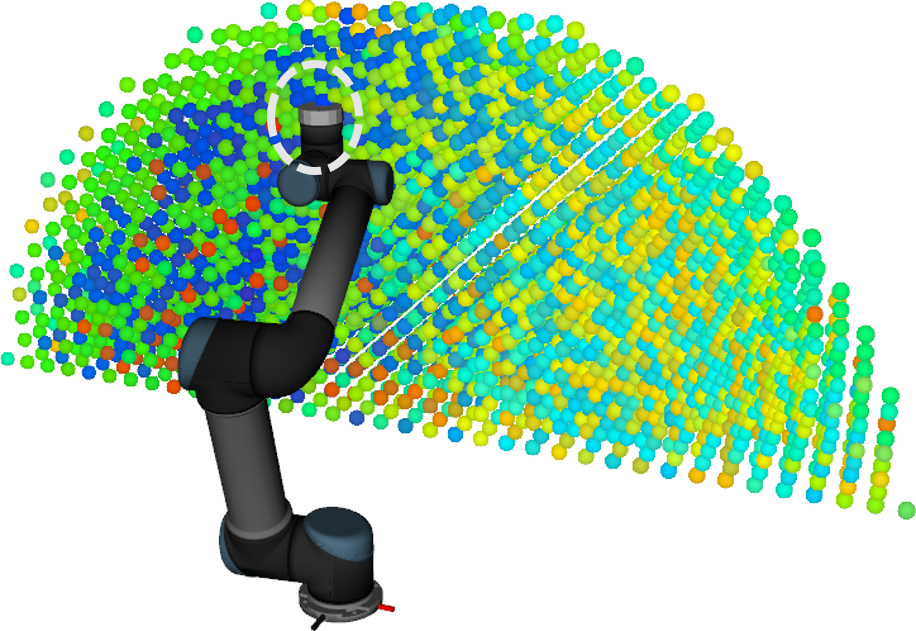}
\caption{Reachability map of our robotic arm, where different colors represent different levels of reachability (i.e., from worst to best in colors Red, Yellow, Green, SkyBlue and Blue). The fixed FDM extruder is placed at the center of a region with high reachability (i.e., the region circled in dash lines). %
} \label{figReachability}
\end{figure}

\begin{figure*}[t]
\includegraphics[width=\linewidth]{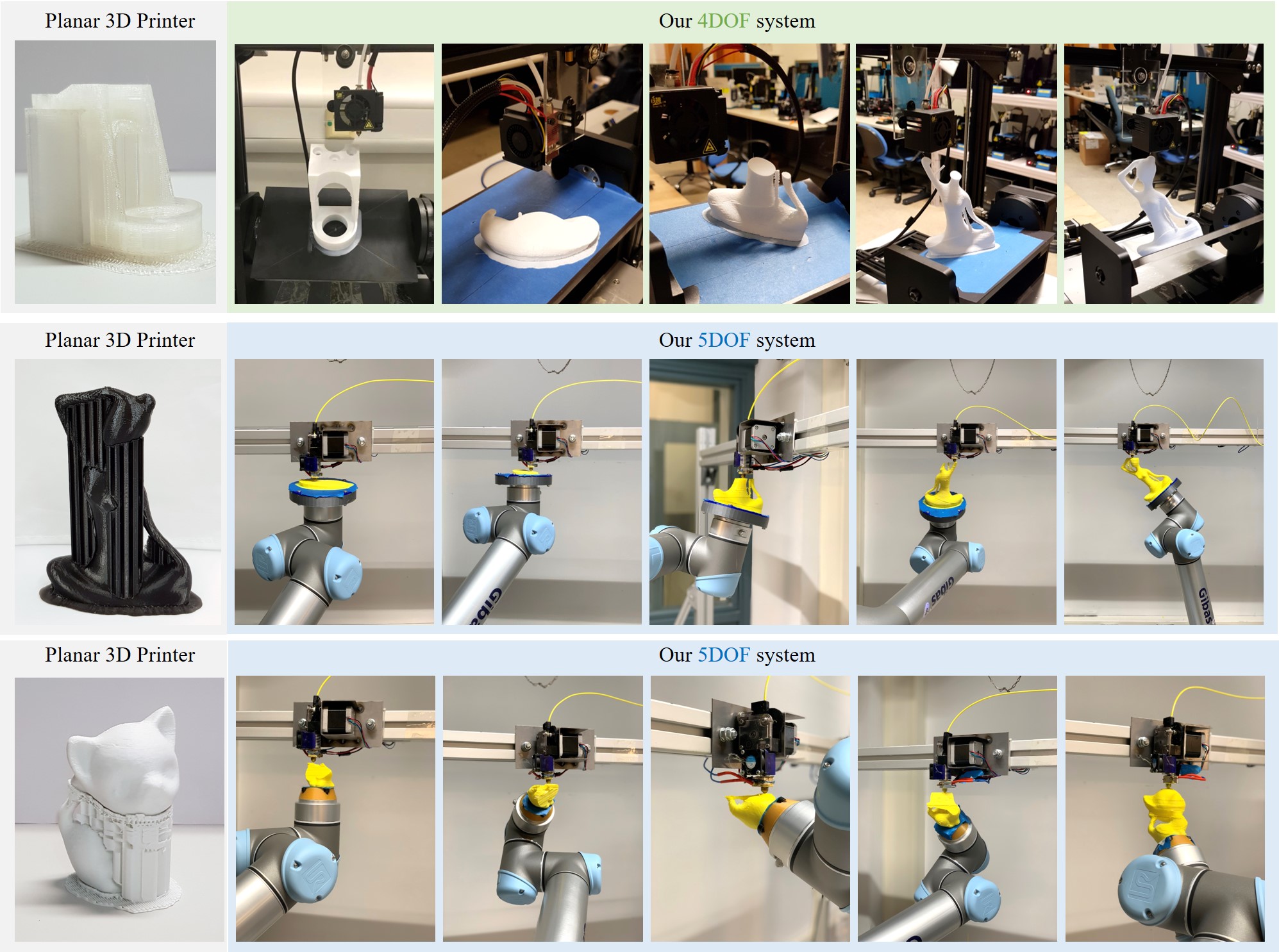}
\caption{The progressive results of fabricating models with our 4DOF multi-directional 3D printing system (the top row) and a 5DOF system realized on a robotic arm (the middle and bottom rows).
}
\label{figPrintingProcess}
\end{figure*}

\subsection{Cartesian-space-based hardware}\label{subsecCartesianSystem}
The hardware setup of the 4DOF multi-directional printer is developed on top of an off-the-shelf FDM printer (i.e., Creality CR-10S and Ultimaker 2+) by adding a rotational platform. Inspired by the 4-axis CNC machine, a turbine-shaft structure driven by a step-motor is built vertically and fixed into the base of the 3D printer platform, which will be moved together. Note that, the additional cost of this hardware system is only about $240$~USD, which is cost-effective when comparing to devices with synchronized multi-axis motion. Moreover, we design an easy-to-calibrate platform, shown in the left of Fig.\ref{figHardwareSystem}, which allows $\pm 60^\circ$ collision-free rotation. %

During the manufacturing process, the motion of the printer header is fully controlled by the 3D printer itself. The newly added motor for rotation is only used to realize the orientation change between sub-models from one to the next. An Arduino chipboard controls the rotation applied to the platform of 3D printing. Note that as a step motor is used, only a limited number of orientations can be realized. This manufacturing constraint is considered while generating sample points on the Gaussian sphere for clipping planes. The process of 3D printing on this 4DOF system can be found in the top row of Fig.\ref{figPrintingProcess} as well as the supplementary video.

\begin{figure*}[t]
\includegraphics[width=\linewidth]{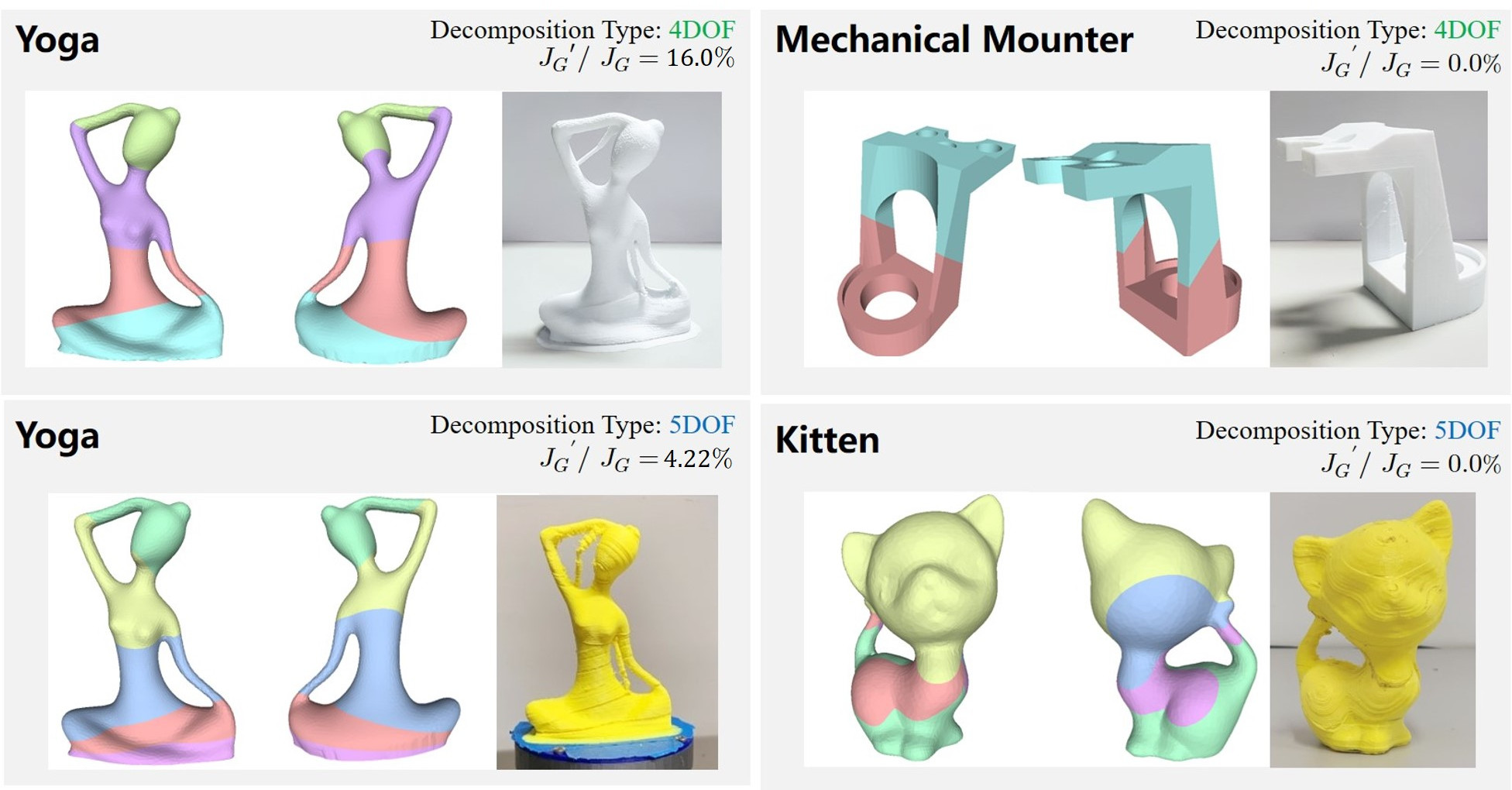}
\caption{The decomposition results fabricated by our system with 4DOF and 5DOF in motion, where the resultant value of $J_G$ is also reported.
}\label{figResult}
\end{figure*}

\begin{figure}[t]
\includegraphics[width=\linewidth]{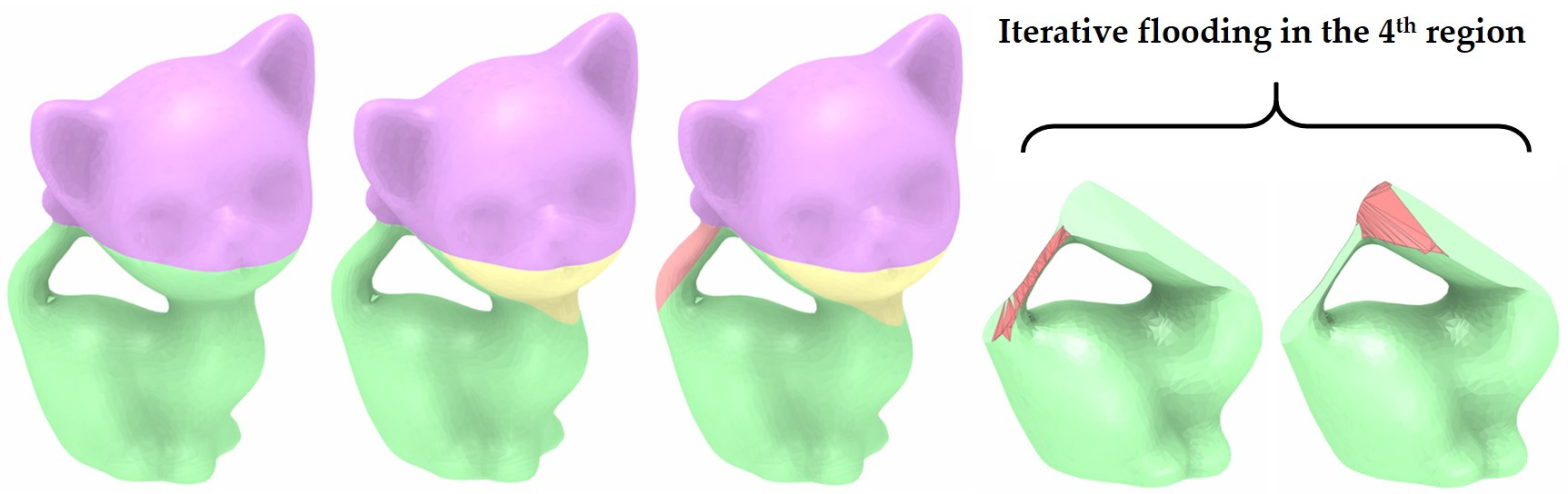}
\caption{
The progressive result of applying the flooding based algorithm \cite{xu18supportfree} to the Kitten model, which is stuck at the fourth region due to handle topology.}\label{figFloodingRes}
\end{figure}

\begin{figure}[t]
\centering
\includegraphics[width=.8\linewidth]{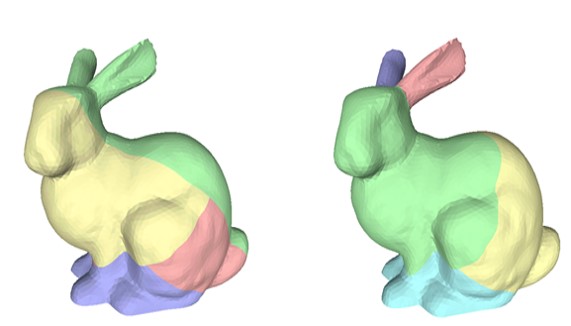}
\caption{
The comparison of our result (left with $J_G=10.85$) and the result of \cite{xu18supportfree} (right with $J_G=31.55$).
}\label{figFloodingCompareRes}
\end{figure}

\subsection{Joint-space-based hardware}
The principle of the 4DOF system can also be extended to 5DOF by using a tilting table with two rotational DOFs. In our experimental tests, a 6DOF robotic arm is used to demonstrate the functionality of our method with 5DOF motion, although a robotic arm provides much lower positioning accuracy. The hardware setup is composed of a UR5 robotic arm, and an FDM extruder fixed on a frame and some other control components. As the extruder is fixed to obtain better adhesion in our system, the change of printing directions and positions is realized by the inverse poses of a printing platform attached to the UR5 robotic arm. Considering the accuracy of positioning that can be achieved on UR5 \cite{Kruit13} and the speed of fabrication, we employ a $0.8$mm diameter nozzle in our system for material deposition.

Because of the hardware constraints of the UR5 robotic arm (e.g., limited ranges of joints), not every point with a given orientation can be realized. The reachability of points inside the working envelope is very sensitive to the relative position of the nozzle in the coordinate system of the UR5's base frame, which needs to be optimized to enhance the reachability. First of all, the workspace of a robotic arm is uniformly sampled into points. For each point in the Cartesian space, we randomly sample an additional $100$ points on the unit sphere around the point with orientations towards the center of the sphere. The reachability map can be generated by Reuleaux~\cite{Reuleaux}. As shown in Fig.\ref{figReachability}, we placed the extruder of our setup at the center of a region with the highest reachability.

The middle and bottom rows of Fig.\ref{figPrintingProcess} present the progressive results of models fabricated on the 5DOF multi-directional 3D printing system. Our method can successfully decompose a given model into support-free components to be fabricated one by one.

\subsection{Results and Discussion}\label{subsecResDiscussion}
We applied our volume decomposition algorithm to a variety of models. In addition to the models shown in Figs.\ref{figTeaser} and \ref{figSchemeComparison}, we tested our system on models with higher genus-number (see Figs.\ref{figPrintingProcess} and \ref{figResult}). Our algorithm can greatly reduce or even eliminate the need for supports on these models. Models that still require supports add these supports to only very small regions of the model, and we compare these results to a conventional planar 3D printer in Fig.\ref{figPrintingProcess}.

The major advantage of our approach compared with \cite{xu18supportfree} is the ability to handle models with handle and loop topology (see the models in Fig.\ref{figResult}). For example, when applying the algorithm of Xu et al.~\cite{xu18supportfree} to the Kitten model of Fig.\ref{figResult}, their flooding algorithm is stuck at the fourth region, as shown in Fig.\ref{figFloodingRes}. When applying their method to the Bunny model with genus-zero topology, the result is similar to ours (see Fig.\ref{figFloodingCompareRes}), though our result has a slightly smaller $J_G$. 
Another problem with their method is that the collision-free constraint has not been explicitly incorporated into the computation -- i.e., the collision between the printer head and the already fabricated ear may happen when printing the other ear of the Bunny.

\begin{table*}[t]
\centering
\caption{Computational Statistic}
\begin{tabular}{c|c|c|c|c|c|c|c|c|c|c|c}
\hline
        &         &  \textbf{Trgl.}  &    &  \textbf{Computing}  &  \textbf{Part}  & \multicolumn{2}{|c|}{$J_G$} & \multicolumn{2}{|c|}{\textbf{Support Volume}} & \multicolumn{2}{c}{\textbf{Printing Time}}         \\
\cline{7-12}
\textbf{Model}   & \textbf{Fig.}    &  \#     & \textbf{Type}   &  \textbf{Time}  &   \#     & \textbf{Before}  &  \textbf{After}  & \emph{Fixed Dir.}  &  \emph{Multi-Dir.} & \emph{Fixed Dir.}  &  \emph{Multi-Dir.}  \\
\hline  \hline
 Bunny        &  \ref{figSchemeComparison}   &  12,420     &    5DOF    &  94 sec.   &  5  &   314.26 &  10.85 &  130.59 &  0.00 & 42 min. &  55 min. \\ 
\hline
Kitten        &  \ref{figSchemeComparison}   &  10,000   &    5DOF    &  98 sec.   & 5   & 1481.36   & 0.81 & 893.58   & 0.00 & 220 min.   & 187 min.\\ 
\hline
Bimba        &  \ref{figSchemeComparison}  &  12,156  &    5DOF    &  105 sec.   &   5   & 560.49   & 20.86   & 382.46   & 14.42  & 108 min.  & 134 min. \\ 
\hline
Fertility        &  \ref{figSchemeComparison}  &  16,172   &    5DOF    &  314 sec.   &   6   & 555.84   & 200.60  & 310.25   & 156.53   & 59 min.   & 70 min. \\ 
\hline
Snowman        &  \ref{figSchemeComparison}   &  10,000      &    5DOF    &   104 sec.  &    5  &  822.34    &  9.98  &  370.92    &  7.33  &  188 min.    &  240 min. \\ 
\cline{2-2} \cline{4-6} \cline{8-8} \cline{10-10} \cline{12-12}
             &    1     &     &    4DOF    &  64 sec. (Total: 192 min.)  &   5  &    & 11.84  &    & 12.62  &    & 242 min.  \\ 
\hline
Yoga        &  \ref{figResult}  &   11,254  &    5DOF    &  298 sec.   &   5    &  613.89  & 25.92    & 392.17  & 39.92    &  189 min.  & 234 min. \\ 
\cline{2-2} \cline{4-6} \cline{8-8} \cline{10-10} \cline{12-12}
                         &  \ref{figResult}  &          &    4DOF    & 88 sec. (Total: 266 min.)   &   4   &    &  98.51 &    & 74.56 &    & 207 min. \\ 
\hline
Mechanical-        &  \ref{figResult} &    15,348      &    5DOF    &  52 sec.   &  2  & 698.09   & 0.00  & 843.45   & 0.00  & 56 min.   & 39 min. \\ 
\cline{2-2} \cline{4-6} \cline{8-8} \cline{10-10} \cline{12-12}
 Mounter     &   \ref{figResult}  &          &    4DOF    &  34 sec. (Total: 102 min.)   &   2    &  & 0.00  &  &  0.00    &  & 39 min. \\ 
\hline
\end{tabular}\label{tabCompStatistic}
\end{table*}

Letting $J_G=0$ is \textit{sufficient} but \textit{unnecessary} for a model $\mathcal{M}$ to be support-freely fabricated by the multi-directional 3D printing system. In other words, when our algorithm returns a decomposition with $J_G \neq 0$ for a model, it is still possible to have a solution for support-free decomposition that was not found. This is partially because of the local optimum determined by the beam-guided search. Our sampling strategy discretizes continuous 3D space, which can also raise this problem.

\begin{figure}[t]
\includegraphics[width=\linewidth]{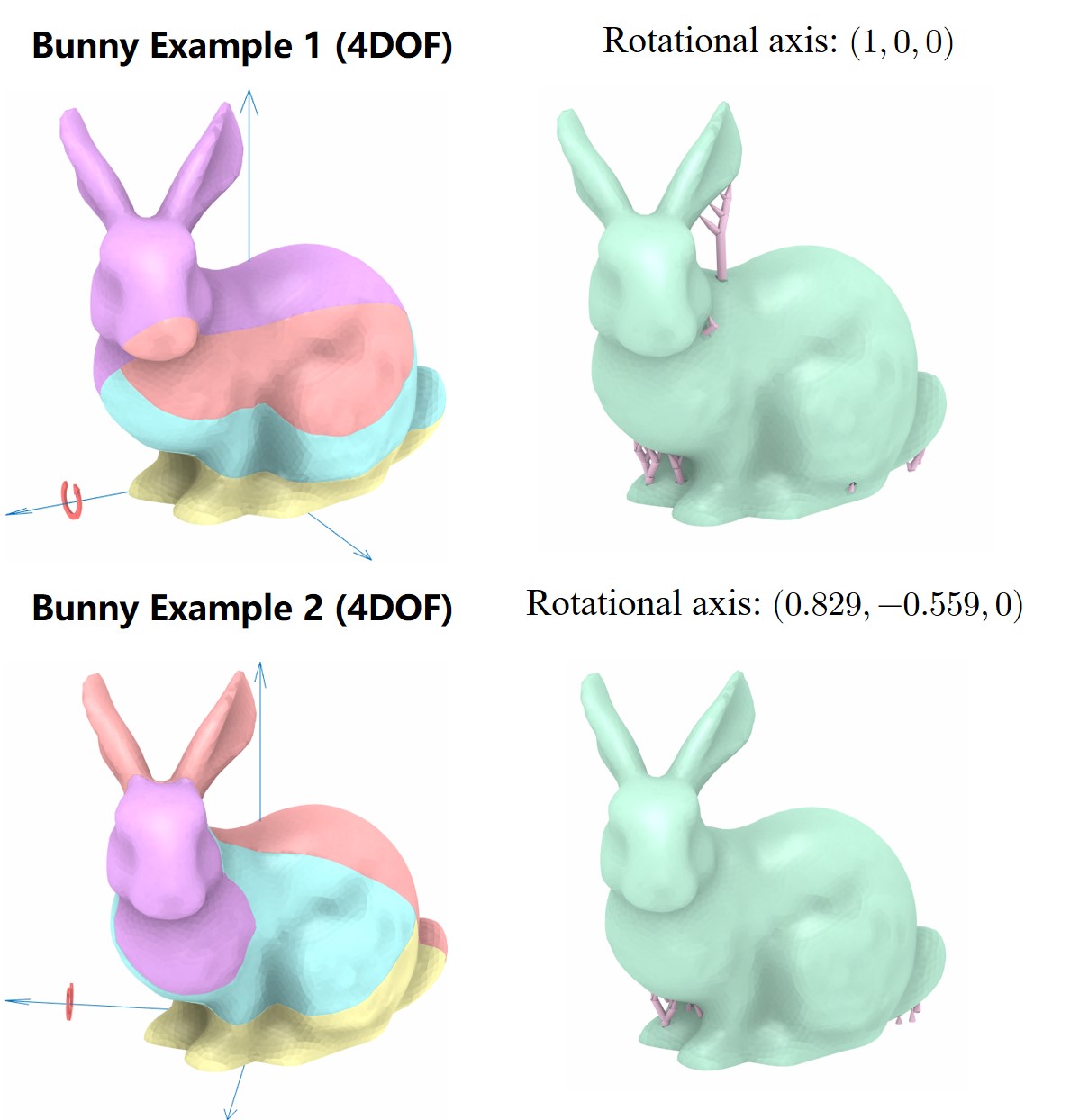}
\caption{
When selecting different axes for 4DOF fabrication, the decomposition gives different results -- (left) $J_G=126.58$ for using $(1,0,0)$ as the rotational axis and (right) $J_G=87.33$ for rotating around $(0.829,-0.559,0)$. The rotational axes are shown in red. 
As can be found in the top row, supporting structures need to be added below the ear of the bunny by the rotational axis $(1,0,0)$. This is eliminated by using $(0.829,-0.559,0)$ as the rotational axis (see the bottom row).
}\label{fig4DOFAxisDir}
\end{figure}

When generating sub-models for 4DOF printing, the resulting decomposition depends on the selection of the rotational axis. As shown in Fig.\ref{fig4DOFAxisDir}, when specifying different axes e.g., $\mathbf{r}_a=(1,0,0)$ and $\mathbf{r}_b=(0.829,-0.559,0)$ as rotational axis, the decomposition results in different levels of self-support. Specifically, we obtain $J_G=126.58$ and $J_G=87.33$ when using $\mathbf{r}_a$ and $\mathbf{r}_b$ as the rotation axis respectively. This brings in a new parameter, the rotational axis, to further optimize the decomposition. A simple solution is to discretely sample a few possible rotational axes, then select the one that leads to the minimal $J_G$ after decomposition.

Our approach employs a sampling strategy for generating candidates of clipping planes, which are then used for computing the decomposition and sequence of multi-directional 3D printing. In one way, this helps us impose manufacturing constraints easily -- e.g., limiting the rotational axis, computing orientations that can be physically realized by step-motors, and excluding the singular and the collided poses for a robotic arm. On the other hand, this also limits the space of computation. The variables in our computations are not continuous, which means we may miss the ``real'' optimal clipping planes. To address this, future work should consider further adjusting clipping planes via continuous optimization by using the planes determined in our approach as an initial guess.

Table~\ref{tabCompStatistic} shows the computational statistics of models tested in this paper. For the 4DOF decomposition, we report both the average time for the beam-search according to a given rotational axis and the total time for searching all possible rotational axes (reported in the bracket), where $180$ possible rotational axes are considered. The computational efficiency of our approach is acceptable when compared with the 3D printing time (i.e., around a few hours in general). We measure printing time using Ultimaker Cura (Version 3.6.0)~\cite{ultimakerCura} with the settings of $0.4$mm layer height and 20\% grid infill. The volumes of support needed for 3D printing along a fix direction (denoted by \emph{Fixed Dir.}) and our method (denoted by \emph{Multi-Dir.}) are reported in Table~\ref{tabCompStatistic}. We also report the comparison of printing time. The reason why longer time is needed for multi-directional 3D printing on some models is the hollowed volume is less on the decomposed components.

Another weakness of this decomposition-based multi-directional 3D printing approach is the relatively weak stiffness of the model. As already studied in \cite{wu2017RoboFDM}, smaller Young's modulus is observed on the specimens generated by this approach during the tensile tests. The weak adhesion of materials mainly causes this at the interface between two regions. One of our future research questions is how to design special structures at the interface between different regions to enhance the mechanical strength of adhesion.

\section{Conclusion}
We present a volume decomposition framework for the support-effective fabrication of general models via multi-directional 3D printing. A beam-guided search computes the decomposition while avoiding local optima. While prior work can only fabricate models with skeletal tree structures, our method can apply to models with multiple loops and handles. We also provide a support generation scheme that allows our framework to fabricate all types of models. The framework can incorporate manufacturing constraints such as the number of rotational axes and the realizable configurations during the orientation sampling process. As a result, our algorithm supports both the 4DOF and the 5DOF systems. We verify the effectiveness of our approach by creating a variety of models on multiple hardware setups.

\section*{Acknowledgment}
This work was partially supported by the seed fund of TU Delft IDE Faculty, 
the Natural Science Foundation of China (61725204, 61521002, 61628211), Royal Society-Newton Advanced Fellowship (NA150431) and MOE-Key Laboratory of Pervasive Computing. The authors would like to thank Nikita Haduong at the University of Washington for polishing the writing.

\ifCLASSOPTIONcaptionsoff
  \newpage
\fi

\bibliographystyle{IEEEtran}%
\bibliography{main}

\begin{thebibliography}{10}
\providecommand{\url}[1]{#1}
\csname url@rmstyle\endcsname
\providecommand{\newblock}{\relax}
\providecommand{\bibinfo}[2]{#2}
\providecommand\BIBentrySTDinterwordspacing{\spaceskip=0pt\relax}
\providecommand\BIBentryALTinterwordstretchfactor{4}
\providecommand\BIBentryALTinterwordspacing{\spaceskip=\fontdimen2\font plus
\BIBentryALTinterwordstretchfactor\fontdimen3\font minus
  \fontdimen4\font\relax}
\providecommand\BIBforeignlanguage[2]{{%
\expandafter\ifx\csname l@#1\endcsname\relax
\typeout{** WARNING: IEEEtran.bst: No hyphenation pattern has been}%
\typeout{** loaded for the language `#1'. Using the pattern for}%
\typeout{** the default language instead.}%
\else
\language=\csname l@#1\endcsname
\fi
#2}}

\bibitem{Gao2015}
W.~Gao, Y.~Zhang, D.~Ramanujan, K.~Ramani, Y.~Chen, C.~B. Williams, C.~C. Wang,
  Y.~C. Shin, S.~Zhang, and P.~D. Zavattieri, ``The status, challenges, and
  future of additive manufacturing in engineering,'' \emph{Computer-Aided
  Design}, vol.~69, pp. 65--89, 2015.

\bibitem{williams1992three}
E.~Sachs, M.~Cima, P.~Williams, D.~Brancazio, and J.~Cornie, ``Three
  dimensional printing: rapid tooling and prototypes directly from a {CAD}
  model,'' \emph{Journal of Engineering for Industry}, vol. 114, no.~4, pp.
  481--488, 1992.

\bibitem{dumas2014bridging}
J.~Dumas, J.~Hergel, and S.~Lefebvre, ``Bridging the gap: Automated steady
  scaffoldings for 3d printing,'' \emph{ACM Trans. Graph.}, vol.~33, no.~4, pp.
  98:1--98:10, July 2014.

\bibitem{vanek2014clever}
J.~Vanek, J.~A. Galicia, and B.~Benes, ``Clever support: Efficient support
  structure generation for digital fabrication,'' in \emph{Computer Graphics
  Forum}, vol.~33, no.~5.\hskip 1em plus 0.5em minus 0.4em\relax Wiley Online
  Library, 2014, pp. 117--125.

\bibitem{zhang2015perceptual}
X.~Zhang, X.~Le, A.~Panotopoulou, E.~Whiting, and C.~C. Wang, ``Perceptual
  models of preference in {3D} printing direction,'' \emph{ACM Trans. Graph.},
  vol.~34, no.~6, p. 215, 2015.

\bibitem{Hu2015a}
K.~Hu, S.~Jin, and C.~C.~L. Wang, ``Support slimming for single material based
  additive manufacturing,'' \emph{Computer-Aided Design}, vol.~65, pp. 1--10,
  2015.

\bibitem{Herholz2015}
P.~Herholz, W.~Matusik, and M.~Alexa, ``Approximating free-form geometry with
  height fields for manufacturing,'' \emph{Computer Graphics Forum}, vol.~34,
  no.~2, pp. 239--251, 2015.

\bibitem{muntoni2018heightblock}
A.~Muntoni, M.~Livesu, R.~Scateni, A.~Sheffer, and D.~Panozzo, ``Axis-aligned
  height-field block decomposition of {3D} shapes,'' \emph{ACM Trans. Graph.},
  2018.

\bibitem{Hu2014}
R.~Hu, H.~Li, H.~Zhang, and D.~Cohen-Or, ``Approximate pyramidal shape
  decomposition,'' \emph{ACM Trans. Graph.}, vol.~33, no.~6, pp. 213:1--213:12,
  2014.

\bibitem{luo2012chopper}
L.~Luo, I.~Baran, S.~Rusinkiewicz, and W.~Matusik, ``Chopper: Partitioning
  models into {3D}-printable parts,'' \emph{ACM Trans. Graph.}, vol.~31, no.~6,
  pp. 129:1--129:9, Nov. 2012.

\bibitem{vanek2014packmerger}
J.~Vanek, J.~Galicia, B.~Benes, R.~M{\v{e}}ch, N.~Carr, O.~Stava, and
  G.~Miller, ``Packmerger: A {3D} print volume optimizer,'' in \emph{Computer
  Graphics Forum}, vol.~33, no.~6.\hskip 1em plus 0.5em minus 0.4em\relax Wiley
  Online Library, 2014, pp. 322--332.

\bibitem{Gao2015UIST}
W.~Gao, Y.~Zhang, D.~C. Nazzetta, K.~Ramani, and R.~J. Cipra, ``{RevoMaker}:
  Enabling multi-directional and functionally-embedded {3D} printing using a
  rotational cuboidal platform,'' in \emph{Proceedings of the 28th Annual ACM
  Symposium on User Interface Software and Technology}, 2015, pp. 437--446.

\bibitem{yao2015level}
M.~Yao, Z.~Chen, L.~Luo, R.~Wang, and H.~Wang, ``Level-set-based partitioning
  and packing optimization of a printable model,'' \emph{ACM Trans. Graph.},
  vol.~34, no.~6, p. 214, 2015.

\bibitem{Chen2015dapper}
X.~Chen, H.~Zhang, J.~Lin, R.~Hu, L.~Lu, Q.~Huang, B.~Benes, D.~Cohen-Or, and
  B.~Chen, ``Dapper: Decompose-and-pack for 3d printing,'' \emph{ACM Trans.
  Graph.}, vol.~34, no.~6, pp. 213:1--213:12, Oct. 2015.

\bibitem{wang2016improved}
W.~Wang, C.~Zanni, and L.~Kobbelt, ``Improved surface quality in {3D} printing
  by optimizing the printing direction,'' in \emph{Computer Graphics Forum},
  vol.~35, no.~2.\hskip 1em plus 0.5em minus 0.4em\relax Wiley Online Library,
  2016, pp. 59--70.

\bibitem{song2016cofifab}
P.~Song, B.~Deng, Z.~Wang, Z.~Dong, W.~Li, C.-W. Fu, and L.~Liu, ``Cofifab:
  coarse-to-fine fabrication of large {3D} objects,'' \emph{ACM Trans. Graph.},
  vol.~35, no.~4, p.~45, 2016.

\bibitem{wei18supportfree}
X.~Wei, S.~Qiu, L.~Zhu, R.~Feng, Y.~Tian, J.~Xi, and Y.~Zheng, ``Toward
  support-free {3D} printing: A skeletal approach for partitioning models,''
  \emph{IEEE Transactions on Visualization and Computer Graphics}, vol.~24,
  no.~10, pp. 2799--2812, Oct 2018.

\bibitem{Branching}
R.~Schmidt and N.~Umetani, ``Branching support structures for {3D} printing,''
  in \emph{ACM SIGGRAPH 2014 Studio}, ser. SIGGRAPH '14.\hskip 1em plus 0.5em
  minus 0.4em\relax New York, NY, USA: ACM, 2014, pp. 9:1--9:1.

\bibitem{Keating2013}
S.~Keating and N.~Oxman, ``Compound fabrication: A multi-functional robotic
  platform for digital design and fabrication,'' \emph{Robotics and
  Computer-Integrated Manufacturing}, vol.~29, no.~6, pp. 439--448, 2013.

\bibitem{Pan2014}
Y.~Pan, C.~Zhou, Y.~Chen, and J.~Partanen, ``Multitool and multi-axis computer
  numerically controlled accumulation for fabricating conformal features on
  curved surfaces,'' \emph{{ASME} Journal of Manufacturing Science and
  Engineering}, vol. 136, no.~3, 2014.

\bibitem{Song2015}
X.~Song, Y.~Pan, and Y.~Chen, ``Development of a low-cost parallel kinematic
  machine for multidirectional additive manufacturing,'' \emph{{ASME} Journal
  of Manufacturing Science and Engineering}, vol. 137, no.~2, 2015.

\bibitem{peng2016fly}
H.~Peng, R.~Wu, S.~Marschner, and F.~Guimbreti{\`e}re, ``On-the-fly print:
  Incremental printing while modelling,'' in \emph{Proceedings of the 2016 CHI
  Conference on Human Factors in Computing Systems}.\hskip 1em plus 0.5em minus
  0.4em\relax ACM, 2016, pp. 887--896.

\bibitem{wu2016printing}
R.~Wu, H.~Peng, F.~Guimbreti{\`e}re, and S.~Marschner, ``Printing arbitrary
  meshes with a 5dof wireframe printer,'' \emph{ACM Trans. Graph.}, vol.~35,
  no.~4, p. 101, 2016.

\bibitem{huang2016framefab}
Y.~Huang, J.~Zhang, X.~Hu, G.~Song, Z.~Liu, L.~Yu, and L.~Liu, ``Framefab:
  robotic fabrication of frame shapes,'' \emph{ACM Trans. Graph.}, vol.~35,
  no.~6, p. 224, 2016.

\bibitem{dai2018support}
C.~Dai, C.~C.~L. Wang, C.~Wu, S.~Lefebvre, G.~Fang, and Y.-J. Liu,
  ``Support-free volume printing by multi-axis motion,'' \emph{ACM Trans.
  Graph.}, vol.~37, no.~4, pp. 134:1--134:14, July 2018.

\bibitem{shembekar2018trajectory}
A.~V. Shembekar, Y.~J. Yoon, A.~Kanyuck, and S.~K. Gupta, ``Generating robot
  trajectories for conformal three-dimensional printing using nonplanar
  layers,'' \emph{Journal of Computing and Information Science in Engineering},
  vol.~19, no.~3, p. 031011, 2019.

\bibitem{xu18supportfree}
K.~{Xu}, L.~{Chen}, and K.~{Tang}, ``Support-free layered process planning
  toward 3 + 2-axis additive manufacturing,'' \emph{IEEE Transactions on
  Automation Science and Engineering}, vol.~16, no.~2, pp. 838--850, April
  2019.

\bibitem{wu2017RoboFDM}
C.~Wu, C.~Dai, G.~Fang, Y.~J. Liu, and C.~C.~L. Wang, ``{RoboFDM}: A robotic
  system for support-free fabrication using {FDM},'' in \emph{2017 IEEE
  International Conference on Robotics and Automation (ICRA)}, May 2017, pp.
  1175--1180.

\bibitem{Lowerre1976}
B.~T. Lowerre, ``The harpy speech recognition system.'' Ph.D. dissertation,
  Carnegie Mellon University, Pittsburgh, PA, USA, 1976, aAI7619331.

\bibitem{huang2014Image}
P.~Huang, C.~C.~L. Wang, and Y.~Chen, ``Algorithms for layered manufacturing in
  image space,'' in \emph{ASME Advances in Computers and Information in
  Engineering Research}, 2014, pp. 377--410.

\bibitem{ultimakerCura}
\BIBentryALTinterwordspacing
Ultimaker. {Ultimaker Cura: Advanced 3D printing software, made accessible}.
  [Online]. Available:
  \url{https://ultimaker.com/en/products/ultimaker-cura-software}
\BIBentrySTDinterwordspacing

\bibitem{Kruit13}
C.J.Kruit, ``A novel additive manufacturing approach using a multiple degrees
  of freedom robotic arm,'' Master's thesis, Delft University of Technology,
  August 2013.

\bibitem{Reuleaux}
A.~Makhal and A.~K. Goins, ``Reuleaux: Robot base placement by reachability
  analysis,'' in \emph{2018 Second IEEE International Conference on Robotic
  Computing (IRC)}, Jan 2018, pp. 137--142.

\end{thebibliography}

\enlargethispage{-6.5cm}

\begin{IEEEbiography}[{\includegraphics[width=1in,height=1.25in,clip,keepaspectratio]{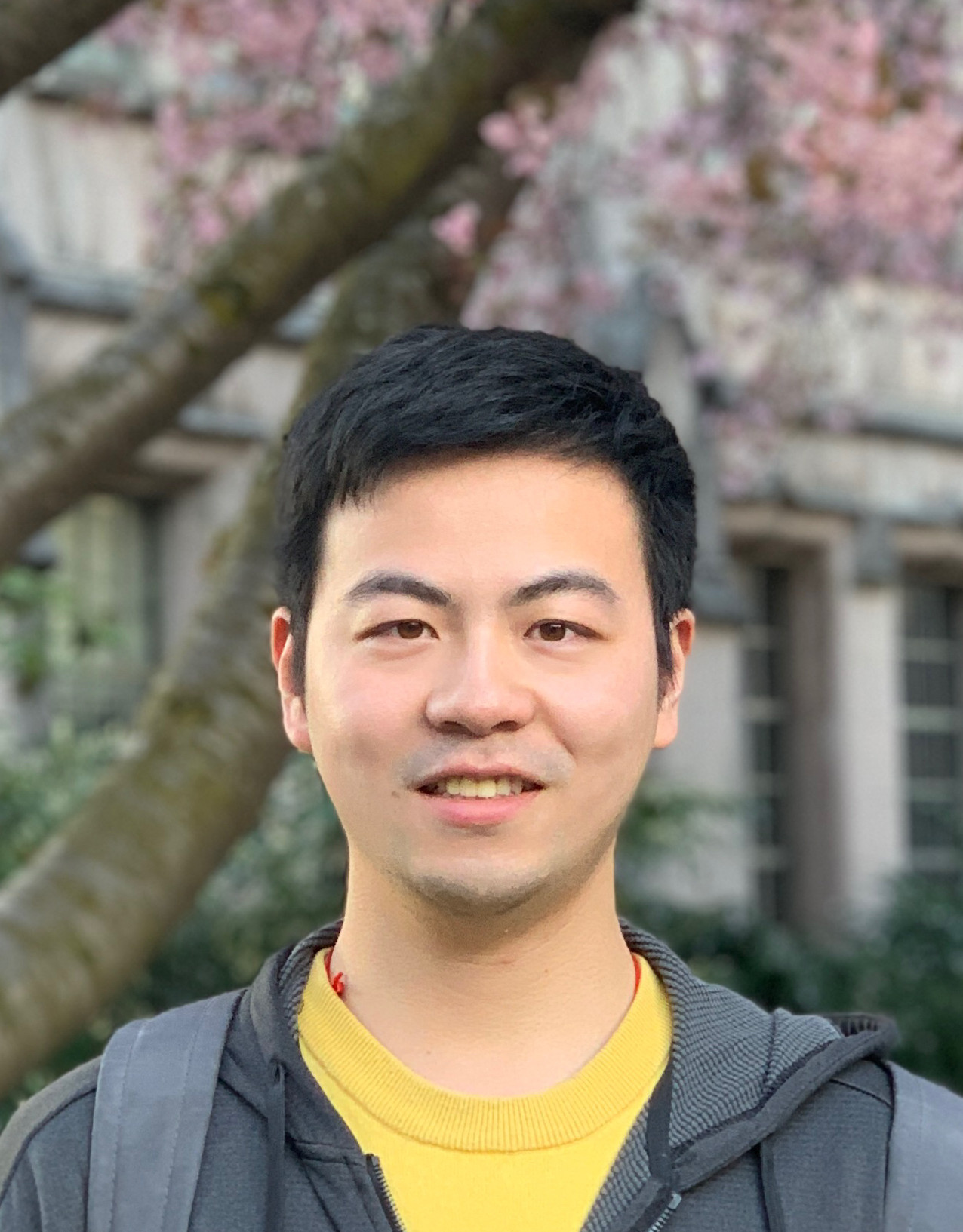}}]{Chenming Wu}
received the B.Eng. degree in electronic information engineering from the
Beijing University of Technology, China,
in 2015. He is currently a Ph.D. candidate
in the Department of Computer Science and Technology, Tsinghua University, China.
His current research interests are intelligent design, computational fabrication and robotics.
\end{IEEEbiography}

\begin{IEEEbiography}[{\includegraphics[width=1in,height=1.25in,clip,keepaspectratio]{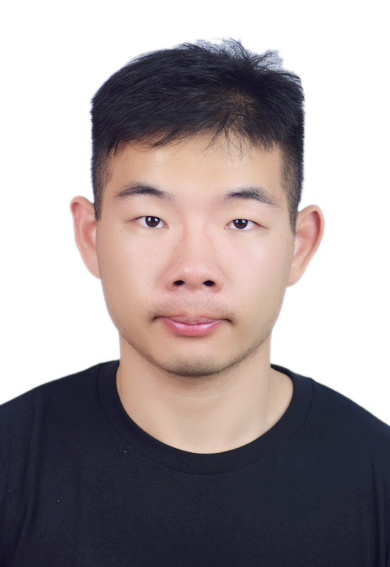}}]{Chengkai Dai}
is currently a Ph.D. candidate of the Department of Design Engineering at Delft University of Technology. His research area includes robotics, geometry computing and computational design.
\end{IEEEbiography}

\begin{IEEEbiography}[{\includegraphics[width=1in,height=1.25in,clip,keepaspectratio]{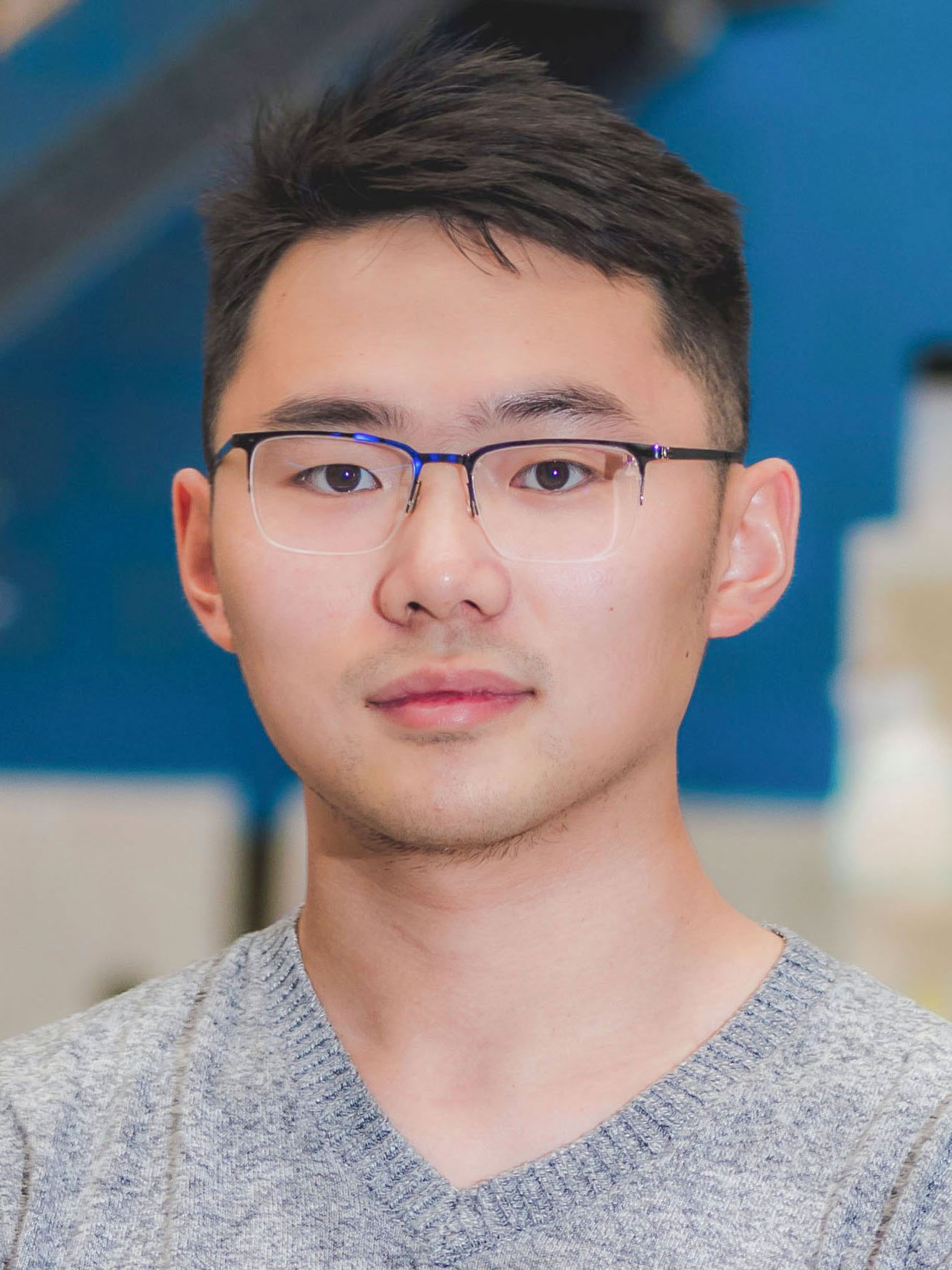}}]{Guoxin Fang}
received the B.Eng. degree in mechanical engineering from the Beijing Institute of Technology, Beijing, China, in 2016. He is currently pursuing the Ph.D. degree with the Department of Design Engineering at Delft University of Technology. His research area includes advanced manufacturing, computational design and robotics.
\end{IEEEbiography}

\begin{IEEEbiography}[{\includegraphics[width=1in,height=1.25in,clip,keepaspectratio]{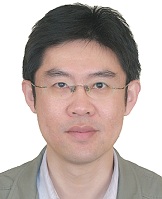}}]{Yong-Jin Liu}
is currently a professor with the Department of Computer
Science and Technology, Tsinghua University,
China. He received his B.Eng degree from Tianjin University, China, in 1998, and the PhD degree from the Hong Kong University of Science
and Technology, Hong Kong, China, in 2004.
His research interests include computational geometry, computer graphics and computer-aided
design. He is a senior member of the IEEE and
a member of ACM.
\end{IEEEbiography}

\begin{IEEEbiography}[{\includegraphics[width=1in,height=1.25in,clip,keepaspectratio]{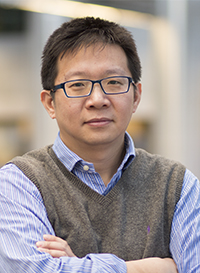}}]{Charlie C.L. Wang}
is currently a Professor of Mechanical and Automation Engineering and Director of Intelligent Design and Manufacturing Institute at the Chinese University of Hong Kong (CUHK). Before that, he was a tenured Professor and Chair of Advanced Manufacturing at Delft University of Technology (TU Delft), The Netherlands. He received the Ph.D. degree (2002) from Hong Kong University of Science and Technology in mechanical engineering, and is now a Fellow of the American Society of Mechanical Engineers (ASME) and the Hong Kong Institute of Engineers (HKIE). His research areas include geometric computing, intelligent design and advanced manufacturing.
\end{IEEEbiography}

\vfill

\end{document}